\documentclass[11pt]{article}

\usepackage[T1]{fontenc}
\usepackage{amsmath}
\usepackage{newtxtext,newtxmath}
\usepackage[letterpaper,top=0.88in,bottom=0.90in,
            left=1in,right=1in,headheight=22pt]{geometry}
\usepackage{booktabs}
\usepackage{graphicx}
\usepackage[table]{xcolor}
\usepackage{colortbl}
\usepackage{array}
\usepackage{makecell}
\usepackage{enumitem}
\usepackage{microtype}
\usepackage{float}
\usepackage{setspace}
\usepackage{abstract}
\usepackage{titlesec}
\usepackage{fancyhdr}
\usepackage{caption}
\usepackage{subcaption}
\usepackage{xurl}
\usepackage{etoolbox}
\usepackage[section]{placeins}
\definecolor{papergray}{HTML}{555555}
\usepackage[authoryear,round]{natbib}
\setcitestyle{aysep={,},yysep={;},notesep={; }}
\usepackage[hidelinks]{hyperref}
\usepackage{tikz}
\usetikzlibrary{arrows.meta,positioning,shapes.geometric,calc,decorations.pathreplacing,fit,backgrounds}

\hypersetup{
  pdftitle={NoLBERT-Calls: Reading Earnings Calls Without Hindsight},
  pdfauthor={Kevin Foley, Jonathan Hartadi, Shivesh Prakash, Hino Wirachapong Suwanphibun, Swapnil Vatsal},
  pdfsubject={A point-in-time encoder adapted to earnings-call transcripts},
  pdfkeywords={earnings calls, look-ahead bias, point-in-time language models, domain adaptation, contrastive learning, answer retrieval}
}

\setlist{nosep,leftmargin=1.5em}

\titleformat{\section}[hang]
  {\Large\bfseries\color{black}}{\thesection}{0.7em}{}
\titleformat{\subsection}[hang]
  {\large\bfseries\color{black}}{\thesubsection}{0.65em}{}
\titleformat{\subsubsection}[hang]
  {\normalsize\bfseries\color{black}}{\thesubsubsection}{0.6em}{}
\titleformat{\paragraph}[runin]
  {\normalsize\bfseries\color{black}}{}{0pt}{}[.]
\titlespacing*{\section}{0pt}{2.2ex plus 0.6ex minus 0.2ex}{1.0ex}
\titlespacing*{\subsection}{0pt}{1.8ex plus 0.4ex minus 0.2ex}{0.7ex}
\titlespacing*{\paragraph}{0pt}{1.0ex plus 0.3ex minus 0.2ex}{0.6em}

\definecolor{navyblue}{RGB}{20, 55, 90}
\definecolor{midblue}{RGB}{40, 90, 135}
\definecolor{softblue}{RGB}{236, 243, 250}
\definecolor{accentteal}{RGB}{30, 120, 130}
\definecolor{softteal}{RGB}{235, 246, 247}
\definecolor{accentorange}{RGB}{190, 80, 25}
\definecolor{softorange}{RGB}{253, 244, 238}
\definecolor{darkgray}{RGB}{40, 45, 50}
\definecolor{midgray}{RGB}{100, 110, 120}
\definecolor{lightgray}{RGB}{246, 248, 250}
\definecolor{cardbg}{RGB}{255, 255, 255}
\definecolor{bordergray}{RGB}{205, 205, 205}

\fancypagestyle{paperstyle}{%
  \fancyhf{}
  \fancyhead[L]{\footnotesize\color{papergray}\textsc{NoLBERT-Calls}}
  \fancyhead[R]{\footnotesize\color{papergray}\thepage}
  \renewcommand{\headrulewidth}{0.35pt}
  \renewcommand{\headrule}{\hbox to\headwidth{\color{bordergray}\leaders\hrule height \headrulewidth\hfill}}
  
}
\AtBeginEnvironment{thebibliography}{\small\setlength{\bibsep}{0.45em}}

\newcommand{\topA}[1]{\cellcolor{accentteal!42}\textbf{#1}}
\newcommand{\topB}[1]{\cellcolor{accentteal!24}#1}
\newcommand{\topC}[1]{\cellcolor{accentteal!10}#1}
\newcommand{\shadekey}{\setlength{\fboxsep}{1.2pt}%
  Shading marks the best (\colorbox{accentteal!42}{\strut\textbf{darkest}}),
  second (\colorbox{accentteal!24}{\strut second}) and third
  (\colorbox{accentteal!10}{\strut third}) value of each column}
\newcolumntype{L}[1]{>{\raggedright\arraybackslash}p{#1}}
\newcommand{\tablenotes}[1]{%
  \par\vspace{0.45em}
  \begin{minipage}{\linewidth}\footnotesize\textit{Notes:} #1\end{minipage}}



\newcommand{\orcidicon}{%
  \href{https://orcid.org/0009-0007-4120-0921}{%
    \raisebox{-0.8pt}{\includegraphics[height=9pt]{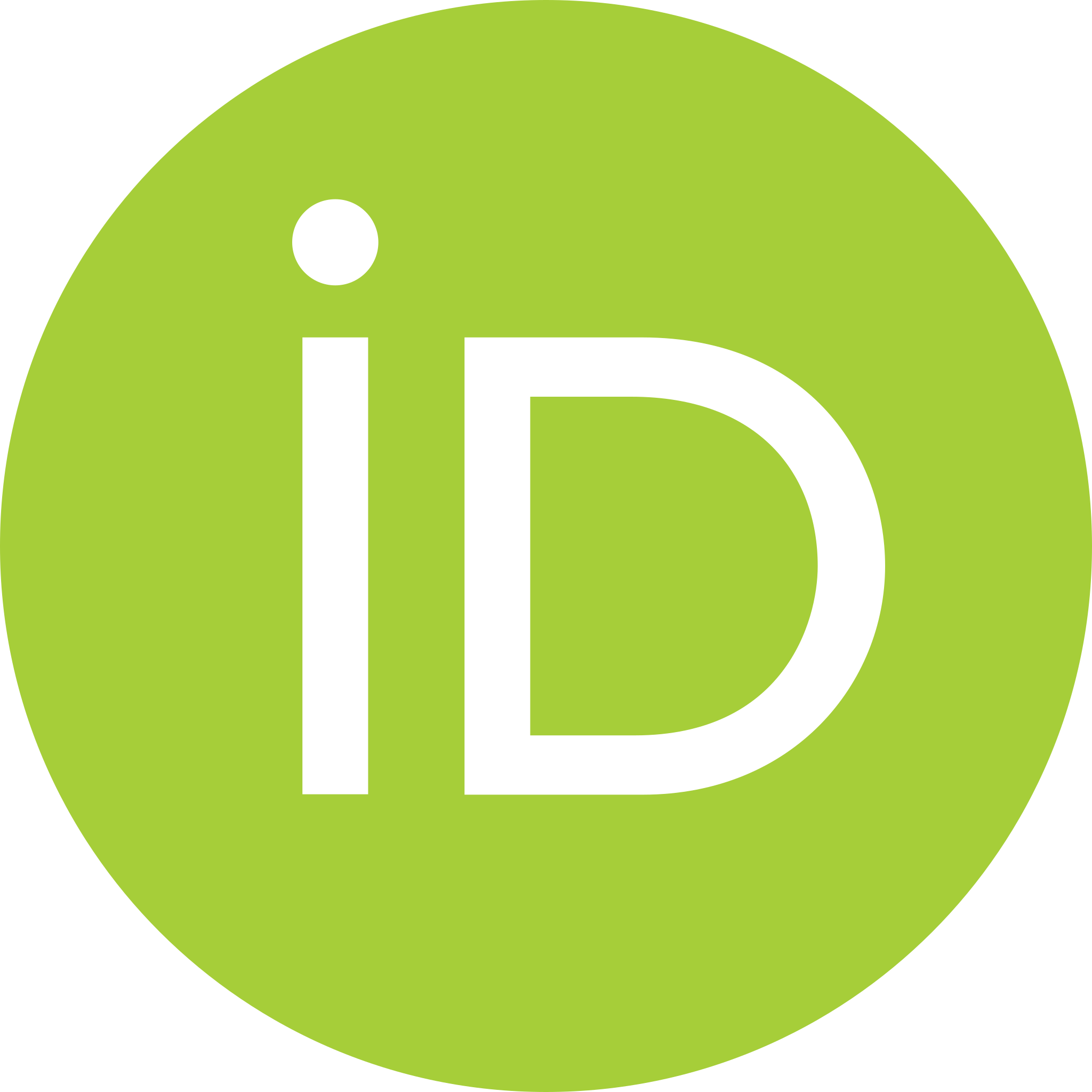}}}}
\renewcommand{\maketitle}{%
  \thispagestyle{empty}
  \vspace*{0.42in}
  \begin{center}
    {\fontsize{24}{29}\selectfont\bfseries
      NoLBERT-Calls:\\[0.12em]
      Reading Earnings Calls Without Hindsight\par}
    \vspace{1.35em}
    {\large
      Kevin Foley\textsuperscript{*}\quad
      Jonathan Hartadi\textsuperscript{*}\\[0.35em]
      Shivesh Prakash\textsuperscript{*,\(\dagger\)}\,\orcidicon\quad
      Hino Wirachapong Suwanphibun\textsuperscript{*}\quad
      Swapnil Vatsal\textsuperscript{*}\par}
    \vspace{0.75em}
    {\normalsize University of California, Berkeley\par}
    {\small Berkeley, California, USA\par}
    \vspace{0.85em}
    {\small
      \href{mailto:kevinfoley3@berkeley.edu}{kevinfoley3@berkeley.edu}
      \enspace\textperiodcentered\enspace
      \href{mailto:johartadi23@berkeley.edu}{johartadi23@berkeley.edu}\\[0.25em]
      \href{mailto:shivesh_prakash@berkeley.edu}{shivesh\_prakash@berkeley.edu}
      \enspace\textperiodcentered\enspace
      \href{mailto:wirachapong.suw@berkeley.edu}{wirachapong.suw@berkeley.edu}
      \enspace\textperiodcentered\enspace
      \href{mailto:swapnil_vatsal@berkeley.edu}{swapnil\_vatsal@berkeley.edu}\par}
  \end{center}
  \vspace{1.45em}
  \begingroup
    \renewcommand{\thefootnote}{\fnsymbol{footnote}}%
    \footnotetext[1]{Equal contribution; authors listed alphabetically by last name. }%
    \footnotetext[2]{Corresponding author.}%
  \endgroup
}

\usepackage{graphicx}
\usepackage{booktabs}
\usepackage{array}
\usepackage{makecell}
\usepackage{adjustbox}
\usepackage{textcomp}
\makeatletter
\@ifundefined{NC@find@L}{\newcolumntype{L}[1]{>{\raggedright\arraybackslash}p{#1}}}{}
\makeatother

\providecommand{\tablenotes}[1]{\par\smallskip\noindent\parbox{\linewidth}{\footnotesize #1}}

\begin{document}
\maketitle

\begin{abstract}
\noindent
Language models are now a standard tool for measuring what managers say on earnings calls, but the encoders behind these measures were usually pretrained on text from the same years a backtest covers. Holding out a test period does not remove that hindsight, because it sits in the features. We adapt NoLBERT, an encoder pretrained only on text published between 1976 and 1995, to earnings calls without letting it read anything dated after 31 December 2023. The calls supply the training signal themselves: every analyst question is followed by management's answer, and the model learns to pick that answer out from the other answers of the same call. We test on every US call from January 2025 to October 2026 (20,323 calls, 254,575 questions) and make every modeling choice on calls from 2024. The adapted model, NoLBERT-Calls-2023, ranks the true answer first for 78.0\% of questions (95\% interval 77.8--78.2; chance 8.0\%), against 26.3\% before adaptation. It is 1.0 point ahead of a call-adapted ChronoBERT, 3.4 points ahead of a call-adapted FinBERT, and 18.7 and 19.8 points ahead of the best off-the-shelf retriever and BM25. A year-by-year test over 2013--2026 finds no look-ahead premium: relative to a model adapted only through 2012, the released model does no better on the years it was trained on than on years it never saw. The price of the cutoff is compute: under one shared 150-minute recipe NoLBERT adapts least well of four encoders, and its lead comes from a longer, staged adaptation of about 13 GPU-hours.
\end{abstract}

\vspace{0.8em}
\begin{center}
\small
\textbf{Keywords:}
earnings calls; look-ahead bias; point-in-time language models; domain adaptation; contrastive learning; answer retrieval
\end{center}
\vfill
\clearpage
\pagestyle{paperstyle}

\section{Introduction}
\label{sec:intro}

Earnings calls are where managers answer analysts in their own words, and a large literature turns those words into measures of disclosure, risk and tone \citep{matsumoto2011calls,hassan2019firmlevel,gow2021nonanswers}. More and more of these measures are built on pretrained language models, and that creates a problem no train--test split can fix. A backtest can keep its test years away from the model it fits, but not from the encoder underneath, which has often read the same years. FinBERT's pretraining corpus includes earnings calls from 2004 to 2019 \citep{huang2023finbert}, and off-the-shelf retrievers such as E5 learn from web text with no control on dates \citep{wang2022e5}. Whatever such an encoder knows about later events is in the features before any downstream model is fitted \citep{kakhbod2025nolbert,sarkar2024lookahead,lopezlira2025memorization}.

Point-in-time encoders remove the problem at its source. NoLBERT is pretrained only on text published between 1976 and 1995 \citep{kakhbod2025nolbert}, and ChronoBERT comes in vintages whose text stops at a chosen year end \citep{he2025chronologically}. Both are general-purpose models, however, and an earnings call is an unusual kind of text. Out of the box they read calls badly. On the task we study, NoLBERT ranks the true answer to an analyst's question first among all answers of the same call for 26.3\% of questions and ChronoBERT-2023 for 24.1\%, while BM25, which only counts shared words \citep{robertson2009bm25}, does so for 58.2\%. A researcher who studies calls has so far had to choose between an encoder that reads them well and one that is free of hindsight.

\begin{figure}[tbp]
\centering
\resizebox{\linewidth}{!}{
\begin{tikzpicture}[
    >=Stealth,
    font=\sffamily,
    panel/.style={draw=bordergray!70, fill=white, rounded corners=3.5pt, line width=0.6pt},
    stepbox/.style={draw=bordergray, fill=lightgray, rounded corners=2pt, line width=0.45pt, text width=2.62cm,
                 minimum height=2.25cm, align=left, inner sep=3.2pt, anchor=north west,
                 font=\fontsize{6.2}{7.8}\selectfont},
    annot/.style={font=\fontsize{6.2}{7.4}\selectfont\color{midgray}},
    seg/.style={font=\bfseries\fontsize{6}{7}\selectfont, color=white},
    dataflow/.style={->, line width=1.0pt, color=navyblue}
]

\draw[panel] (0, 0) rectangle (16.2, 6.75);
\node[anchor=north west, font=\bfseries\scriptsize\color{navyblue}] at (0.15, 6.60)
    {\textsc{From transcripts to a hindsight-free reader}};

\node[stepbox] (s1) at (0.25, 6.05)
    {{\bfseries\fontsize{7.2}{8.4}\selectfont\color{navyblue}1. Transcripts}\\[2pt]
     S\&P Capital IQ\\ 416,463 calls\\ 16,657 firms\\ 2004 to Oct 2026};
\node[stepbox] (s2) at (3.46, 6.05)
    {{\bfseries\fontsize{7.2}{8.4}\selectfont\color{navyblue}2. Exchanges}\\[2pt]
     question $\rightarrow$ answer pairs\\ 200-word passages\\ 3.05M exchanges\\ no labels, no returns};
\node[stepbox, fill=softblue, draw=midblue!60] (s3) at (6.67, 6.05)
    {{\bfseries\fontsize{7.2}{8.4}\selectfont\color{midblue}3. Stage 1}\\[2pt]
     start: NoLBERT\\ calls up to 2012\\ all countries\\ 480 minutes};
\node[stepbox, fill=softteal, draw=accentteal!70] (s4) at (9.88, 6.05)
    {{\bfseries\fontsize{7.2}{8.4}\selectfont\color{accentteal}4. Stage 2}\\[2pt]
     start: stage 1\\ US calls up to 2023\\ 2 $\times$ 150 minutes\\ \textbf{NoLBERT-Calls-2023}};
\node[stepbox, fill=softorange, draw=accentorange!70] (s5) at (13.09, 6.05)
    {{\bfseries\fontsize{7.2}{8.4}\selectfont\color{accentorange}5. Test}\\[2pt]
     answer retrieval\\ within each call\\ 20,323 calls\\ 254,575 questions};
\foreach \a/\b in {s1/s2, s2/s3, s3/s4, s4/s5}
    \draw[dataflow] (\a.east) -- (\b.west);

\begin{scope}[shift={(0, 0.42)}]
  \tikzset{lane/.style={font=\bfseries\fontsize{5.9}{7}\selectfont, color=white},
           lanelab/.style={font=\fontsize{6.1}{7.2}\selectfont, anchor=east}}
  \draw[midgray, line width=0.45pt] (0.5, 0) -- (4.35, 0);
  \draw[midgray, line width=0.45pt] (4.85, 0) -- (15.75, 0);
  \draw[midgray, line width=0.6pt] (4.44, -0.13) -- (4.56, 0.13);
  \draw[midgray, line width=0.6pt] (4.64, -0.13) -- (4.76, 0.13);
  \foreach \x/\y in {0.6/1976, 4.2/1995, 5.0/2006, 8.562/2013, 11.616/2019, 14.160/2024}
     {\draw[midgray, line width=0.4pt] (\x, 0) -- (\x, -0.07);
      \node[annot, anchor=north] at (\x, -0.09) {\y};}
  \draw[midgray, line width=0.4pt] (15.6, 0) -- (15.6, -0.07);
  \node[annot, anchor=north] at (15.6, -0.09) {Oct 2026};

  \fill[navyblue] (0.6, 2.62) rectangle (4.2, 2.94);
  \node[lane] at (2.4, 2.78) {NoLBERT pretraining text};
  \fill[midblue!85] (5.0, 2.12) rectangle (8.542, 2.44);
  \node[lane] at (6.77, 2.28) {stage 1: calls, all countries};
  \fill[accentteal] (5.0, 1.62) rectangle (14.140, 1.94);
  \node[lane] at (9.57, 1.78) {stage 2: US calls up to 31 Dec 2023};
  \fill[accentorange!45] (14.180, 1.12) rectangle (14.649, 1.44);
  \fill[accentorange]    (14.689, 1.12) rectangle (15.6, 1.44);
  \node[lanelab, color=accentorange] at (14.05, 1.28) {choices on 300 calls of 2024; test on every US call of 2025--26};
  \fill[bordergray!45] (8.562, 0.62) rectangle (14.649, 0.94);
  \foreach \k in {0,...,11} {\fill[white] ({8.562 + 0.5089*\k + 0.2545}, 0.78) circle (0.045);}
  \fill[accentorange!30] (14.689, 0.62) rectangle (15.6, 0.94);
  \node[lanelab, color=darkgray] at (8.45, 0.78) {time test: 500 calls a year, then the main test};

  \draw[darkgray, line width=0.7pt, dash pattern=on 2.2pt off 1.6pt] (14.160, 0.45) -- (14.160, 3.30);
  \node[font=\bfseries\fontsize{6.2}{7.4}\selectfont, color=darkgray, anchor=south east] at (14.10, 3.02)
      {cutoff: the released model has read no text after this date};
\end{scope}
\end{tikzpicture}}
\caption{From transcripts to a reader without hindsight. Top: the five steps. Bottom: what each step reads, by date. NoLBERT's pretraining text ends in 1995. Stage~1 reads calls from all countries up to 2012 and stage~2 reads US calls up to 31 December 2023. Every choice is made on 300 calls from 2024, and the test covers every US call from January 2025 to October 2026. The time test scores 500 calls a year from 2013 to 2024 and then the main test.}
\label{fig:pipeline}
\end{figure}
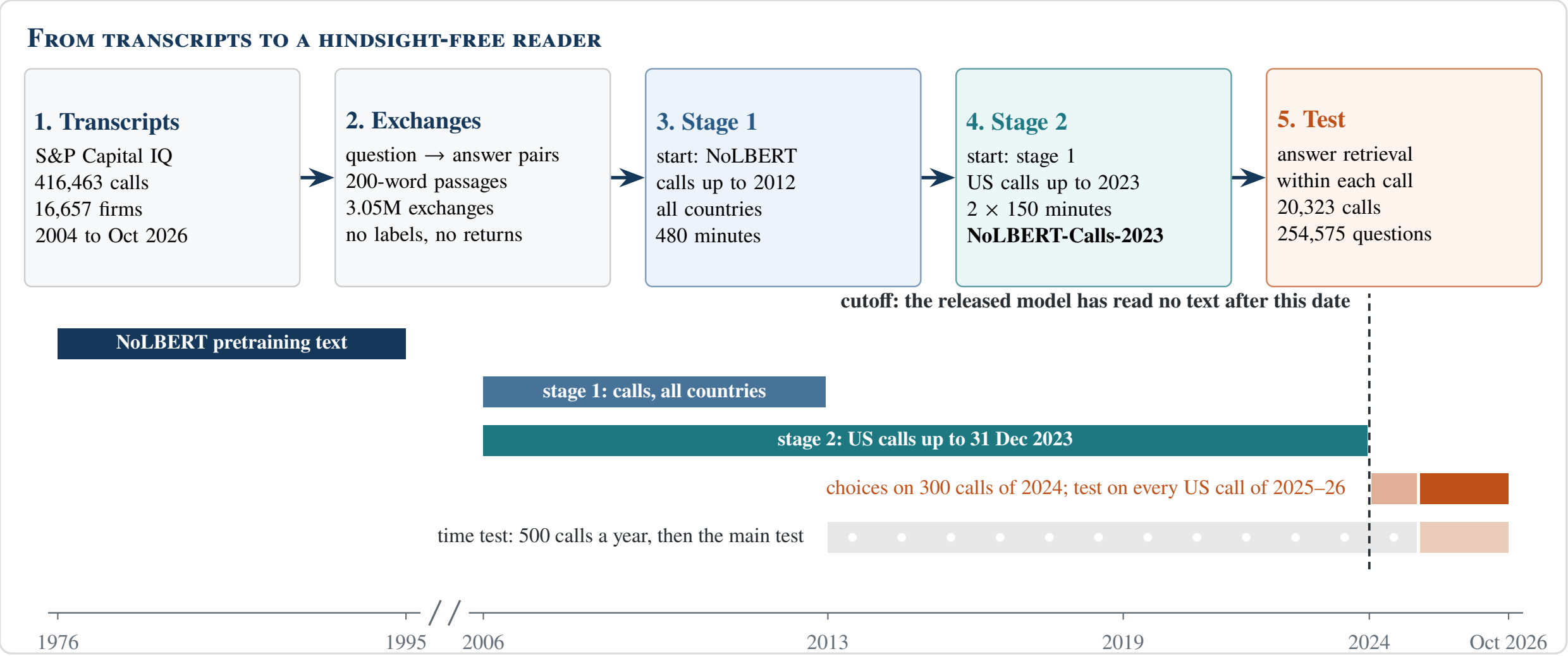

We show that the choice is unnecessary. Calls carry their own supervision: in the question-and-answer session, each analyst question is followed by management's answer. The pairing needs no labels and no returns, and every transcript has it. We adapt NoLBERT to calls in two stages. A contrastive loss asks each question to find its own answer among the other answers of the same call, which share the firm, the quarter and much of the vocabulary, and a masked-word loss teaches the vocabulary of calls. Stage~1 trains on calls from all countries up to 2012 and stage~2 on US calls up to 2023. The released model, NoLBERT-Calls-2023, has read no text dated after 31 December 2023, and the whole adaptation takes about 13 GPU-hours on one consumer GPU and one T4 (Figure~\ref{fig:pipeline}).

The evaluation is built so that hindsight cannot help and the test is used only once. We score every US call from January 2025 to October 2026, 20,323 calls with 254,575 analyst questions, all dated after the training text of every model we compare. Every choice, from the checkpoint and the training recipe to the model we release, was made on 300 calls from 2024. The comparison covers word-overlap baselines, five pretrained encoders used as they are, three off-the-shelf retrievers, and four encoders adapted to calls with one shared recipe and budget. A second test scores 500 calls a year from 2013 to 2024 and asks whether an encoder adapted through 2023 does better on the years it trained on than on the years after its cutoff.

Four results stand out. First, NoLBERT-Calls-2023 ranks the true answer first for 78.0\% of test questions (95\% interval 77.8--78.2) and for 79.0\% when blended with word overlap. It leads a call-adapted ChronoBERT by 1.0 point (0.9--1.2) and a call-adapted FinBERT by 3.4 (3.2--3.5), and it leads the best retriever and BM25 by 18.7 and 19.8 points. The order holds at every call size. Second, how well an encoder reads calls depends on adaptation and hardly at all on the date of its pretraining text. Used zero-shot, every encoder scores below 32\%, and ChronoBERT-2023 does no better than NoLBERT although its text is 28 years more recent; after adaptation, every encoder reaches 67--78\%. Third, we find no look-ahead premium. Against a model adapted only through 2012, the released model leads by 12.0 points on calls from 2013--2023, the years it trained on, and by 12.9 points on calls from 2025--2026, which it never saw; measured against the three years just before its cutoff, the difference is 0.3 points, with a 95\% interval that includes zero. Fourth, the cutoff has a price, and it is paid in compute. Under one shared 150-minute recipe, NoLBERT adapts least well of the four encoders (67.0\% against 74.7\% for FinBERT). A longer, staged adaptation and a longer window on each answer are what put it ahead.

For research on earnings calls, the result is an encoder whose knowledge stops at a known date. Any measure built on it, such as how directly an answer responds to its question, can be backtested from 2024 onward without hindsight, and the recipe is cheap enough to repeat for other cutoffs. Section~\ref{sec:related} reviews related work and Section~\ref{sec:data} describes the data. Section~\ref{sec:methods} sets out the readers, the adaptation and the tests, Section~\ref{sec:results} reports the results, and Section~\ref{sec:discussion} discusses limitations and uses.

\section{Related Work}
\label{sec:related}

\paragraph{Look-ahead bias and point-in-time models}
A language model trained on text from the period of a study can carry later information into the study's features. Large language models recall economic and financial data from their training period \citep{lopezlira2025memorization}, pretrained models used for forecasting show look-ahead bias \citep{sarkar2024lookahead}, and commercial models applied to accounting and finance tasks draw on it \citep{levy2024caution}. \citet{ludwig2025llm} treat the absence of leakage between a model's training text and the research sample as a condition for using the model's output in empirical work. One response hides the firm from the model, either by anonymizing news headlines \citep{glasserman2023lookahead} or by rewriting a text until a model can no longer tell which firm it describes \citep{engelberg2025entity}. On earnings calls, though, anonymization can destroy more information than the bias it removes \citep{wu2025anonymization}. A second response adjusts a large model at inference time \citep{merchant2025fast}. A third trains models only on text up to a cutoff. StoriesLM \citep{sarkar2024storieslm}, ChronoBERT \citep{he2025chronologically} and DatedGPT \citep{yan2026datedgpt} are families of models with annual cutoffs, and NoLBERT \citep{kakhbod2025nolbert} is an encoder pretrained only on text from 1976 to 1995; with such models, \citet{he2025chronologically} find the bias in news-based return prediction to be modest. These are general-purpose models. NoLBERT has been adapted to financial news, with a contrastive loss on passages of the same article, to build topics for the cross-section of stock returns \citep{foley2026readingnewshindsightnolbert,prakash2026readingnewsssrn}, but not to earnings calls. We take the third route for calls and carry a point-in-time encoder into this domain without moving its knowledge past a chosen date.

\paragraph{Domain adaptation and retrieval}
Continued pretraining on text from a target domain improves an encoder on tasks in that domain \citep{gururangan2020dont}. In finance this produced several FinBERT models \citep{araci2019finbert,yang2020finbert,huang2023finbert} and the much larger BloombergGPT \citep{wu2023bloomberggpt}. A second line of work trains encoders to place related texts close together. Sentence-BERT \citep{reimers2019sentencebert}, SimCSE \citep{gao2021simcse} and dense passage retrieval \citep{karpukhin2020dense} use contrastive losses of the InfoNCE type \citep{oord2018cpc}, as do general-purpose retrievers such as E5 \citep{wang2022e5}, BGE \citep{xiao2023bge} and all-mpnet-base-v2, which is built on MPNet \citep{song2020mpnet}. Sentence embeddings also make better topic models for asset pricing: in the narrative framework of \citet{bybee2023narrative}, topics from a sentence transformer beat LDA topics on both coherence and Sharpe ratio \citep{foley2026wordcountscontexttopic,prakash2026wordcountsssrn}. On FinMTEB, embedding models adapted to finance beat general-purpose ones, and scores on general benchmarks say little about scores on financial text \citep{tang2025finmteb}. BM25 \citep{robertson2009bm25} remains the standard word-overlap baseline. We pair a contrastive loss with a masked-word loss and take the training pairs from the calls themselves, so the adaptation needs neither labels nor synthetic data, and it reads nothing from after the cutoff.

\paragraph{Earnings calls in finance and NLP}
Text is now a standard data source in economics and finance \citep{gentzkow2019text}, from Federal Reserve records \citep{kakhbod2026fedmind} and patents \citep{kakhbod2025creative,hochberg2026patents} to social media \citep{kakhbod2025finfluencers} and surveys \citep{fedyk2026ai}. Earnings calls are among the most studied of these texts, and much of their value lies in the question-and-answer session. The discussion with analysts is more informative than the prepared remarks \citep{matsumoto2011calls}, and its tone predicts returns after the call \citep{price2012tone}. What managers decline to say is informative too \citep{hollander2010silence}, which has led to measures of non-answers \citep{gow2021nonanswers,barth2023nonanswers}. Calls are also the basis of firm-level measures of political risk \citep{hassan2019firmlevel}, corporate culture \citep{li2021culture} and exposure to climate change \citep{sautner2023climate}, built with dictionaries \citep{loughran2011liability}, with word lists and embeddings learned from the calls, or with encoders such as FinBERT \citep{huang2023finbert}. In NLP, calls are used to forecast volatility from text and audio \citep{qin2019volatility,li2020maec,yang2020html,sawhney2020voltage}, to benchmark summarization \citep{mukherjee2022ectsum} and to label the qualities of answers \citep{pardawala2024subjectiveqa}. \citet{yu2025dec} find that the transcript representations used to forecast volatility mostly encode which firm is speaking. Our task holds the firm fixed, because every candidate answer comes from the same call. None of these benchmarks asks when a model's training text was written; every call in our test was held after it.

\section{Data}
\label{sec:data}

\subsection{Earnings-Call Transcripts}
\label{sec:data:calls}

We use the earnings-call transcripts of S\&P Capital IQ, obtained through WRDS \citep{ciqtranscripts}, and take every event that Capital IQ labels as an earnings call (key-development type 48). Most calls exist in several versions, from a first draft to an audited copy. We keep the most processed version of each, which reduces 1,318,483 transcripts to 416,463 calls from 16,657 firms in 98 countries. The calls run to 31 October 2026, and almost all of them were held from 2006 on. We count a call as a US call if any of its linked firms is a US firm, which gives 232,324 calls from 7,427 firms. Nearly every call (99.7\%) carries a GVKEY, so the corpus links directly to Compustat and CRSP.

\begin{figure}[tbp]
\centering
\includegraphics[width=\linewidth]{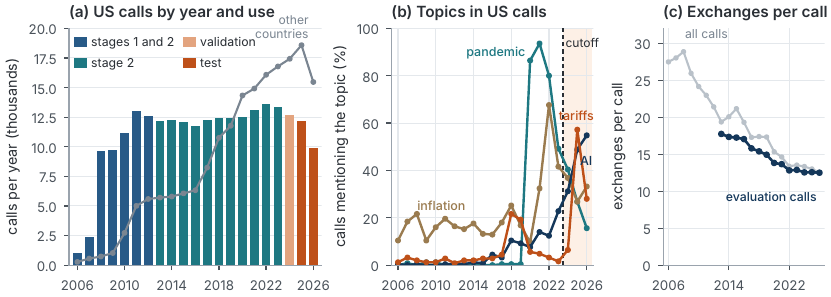}
\caption{US earnings calls. (a) Calls per year and their role in this paper. The line counts calls of firms outside the US, which enter stage~1 up to 2012. (b) Share of calls that mention each subject, in a random sample of 250 US calls a year; the shaded years come after the cutoff of the released model. (c) Analyst exchanges per call, in the same samples (all calls) and in the calls of the time test and the main test (evaluation calls, at least three exchanges each). 2026 runs to 31 October.}
\label{fig:data}
\end{figure}

US coverage is stable over time, at 11,000 to 14,000 calls in every full year since 2010, while coverage of other countries grows from 5,004 calls in 2011 to 18,578 in 2025 (Figure~\ref{fig:data}a). We drop a call if it has no text, if the numbering of its turns has gaps, if it has fewer than 500 words, if it duplicates another call of the same firm, or if it is not in English, which we flag when common English function words make up less than a quarter of its words or when it contains long stretches of non-English text. The transcripts are licensed, and neither this paper nor the released materials reproduce any of their text.

A call opens with prepared remarks by management and continues with questions from analysts. In a random sample of 5,288 US calls, stratified by year, the median call has about 6,900 words, 2,900 of them in the prepared remarks and 3,800 in the question-and-answer session. A median of six analysts ask questions, and executives speak 82\% of the words. The session has changed shape over time. Its length in words has fallen only a little, from a median of about 4,500 words in 2006 to 3,700 in 2025, but the number of exchanges in a call has more than halved, from 27 to 13 (Figure~\ref{fig:data}c), so the typical answer is now much longer. The subjects have changed as well (Figure~\ref{fig:data}b). The pandemic, mentioned in 0.4\% of sampled calls in 2019, appears in 86\% in 2020. Tariffs go from 6\% of calls in 2024 to 57\% in 2025, and artificial intelligence from 23\% in 2023 to 55\% in 2026. The test years thus bring subjects at intensities that no year of the training data reached.

\subsection{Exchanges and Passages}
\label{sec:data:pairs}

We cut each call into two kinds of units. An \emph{exchange} is a question with its answer. In the question-and-answer session, consecutive turns of an analyst form the question, and the turns of management that follow, up to the next analyst question, form the answer; turns of the operator are skipped. A \emph{passage} is a 200-word window of management's speech. US calls up to the end of 2023 give 196,569 calls, 188,692 of them with at least one exchange, 3,052,636 exchanges and 785,924 passages. US calls up to the end of 2012 give 59,523 calls and 1,067,124 exchanges.

\subsection{Sample Splits}
\label{sec:data:splits}

Table~\ref{tab:splits} lists the splits, all fixed before any model was trained. The main protocol has a cutoff of 31 December 2023: models train on US calls up to that date, 300 calls drawn at random from the 12,620 US calls of 2024 serve for validation (3,763 questions), and the test is every US call from 1 January 2025 to 31 October 2026 with at least three exchanges, 20,323 calls and 254,575 questions (11,183 calls in 2025 and 9,140 in 2026). The time protocol has a cutoff of 31 December 2012. It validates on 300 US calls from 2013--2015 (5,398 questions) and tests on 500 random US calls a year from 2013 to 2024, each with at least three exchanges (between 6,268 and 8,864 questions a year), followed by the main test. Stage~1 of our model is the one exception to the US-only training data: it trains on calls from all countries up to 2012 (Section~\ref{sec:methods:finetune}).

Validation and test calls are drawn once with a fixed seed and are the same for every reader. Every choice in this paper is made on the validation calls; the test calls are used only to report results.

\begin{table}[H]
\centering
\footnotesize
\caption{Sample splits.}
\label{tab:splits}
\setlength{\tabcolsep}{4pt}
\begin{tabular}{@{}L{2.6cm}L{4.3cm}L{3.6cm}L{4.3cm}@{}}
\toprule
Protocol & Training calls & Validation & Test \\
\midrule
Main (cutoff 2023) & US calls up to 31 Dec 2023: 196,569 calls, 3,052,636 exchanges, 785,924 passages & 300 US calls from 2024 (3,763 questions) & Every US call from Jan 2025 to Oct 2026: 20,323 calls, 254,575 questions \\
\addlinespace
Time (cutoff 2012) & US calls up to 31 Dec 2012: 59,523 calls, 1,067,124 exchanges & 300 US calls from 2013--2015 (5,398 questions) & 500 US calls a year, 2013--2024 (6,268 to 8,864 questions a year), then the main test \\
 
\bottomrule
\end{tabular}
\tablenotes{All calls are US calls, and test calls have at least three exchanges. Validation and test calls are drawn once with a fixed seed and are the same for every reader. Stage~1 of the released model trains on calls from all countries up to 2012.}
\end{table}

\subsection{Public Benchmarks}
\label{sec:data:benchmarks}

Two public benchmarks test the adapted encoders away from our own task. SubjECTive-QA \citep{pardawala2024subjectiveqa} labels answers from earnings calls on six properties and comes with three train--test splits published by its authors. GLUE \citep{wang2019glue} measures general language understanding; we use the official development sets of CoLA, QNLI and SST-2. Neither benchmark was used for any modeling choice.

\section{Method}
\label{sec:methods}

\subsection{Overview}
\label{sec:methods:overview}

Figure~\ref{fig:pipeline} shows the pipeline and the dates that matter. We cut the cleaned transcripts into exchanges and passages, adapt NoLBERT to calls in two stages, the first on calls up to 2012 and the second on US calls up to 2023, make every choice on 300 calls from 2024, and test on US calls from 2025 and 2026. NoLBERT's pretraining text ends in 1995 and the adaptation adds nothing dated after 31 December 2023, so a study that starts in 2024 or later can use the model without hindsight. We call any method that scores a question against candidate answers a \emph{reader}. All readers see the same calls and the same questions.

\subsection{Readers Compared}
\label{sec:methods:readers}

Table~\ref{tab:readers} lists the readers in four groups, chosen so that the effect of adaptation can be told apart from the effect of the starting model.

\emph{Word overlap.} TF-IDF and BM25 \citep{robertson2009bm25}, with $k_1 = 1.2$ and $b = 0.75$, use word weights fitted on calls up to the cutoff. They need no training and set the bar that any encoder has to clear.

\emph{Pretrained encoders used as they are.} NoLBERT \citep{kakhbod2025nolbert}, ChronoBERT-2012 and ChronoBERT-2023 \citep{he2025chronologically}, FinBERT \citep{huang2023finbert} and DeBERTa-v3-base \citep{he2021debertav3} are read with mean pooling at pinned revisions.

\emph{Off-the-shelf retrievers.} all-mpnet-base-v2 \citep{song2020mpnet,reimers2019sentencebert}, BGE-base-en-v1.5 \citep{xiao2023bge} and E5-base-v2 \citep{wang2022e5} are trained for retrieval on web-scale data and read with their own pooling and prefixes.

\emph{Encoders adapted to calls.} To separate the effect of adaptation from that of the starting model, we adapt NoLBERT, ChronoBERT, FinBERT and DeBERTa-v3 to US calls up to 2023 with one shared recipe: one run of 150 minutes on one T4, up to eight exchanges of a call per batch, and 128 tokens for questions and answers, with the objective of Section~\ref{sec:methods:finetune}. For the time test we also adapt NoLBERT and ChronoBERT to US calls up to 2012 with the same recipe. Our staged model, NoLBERT-Calls-2023, and its first stage complete the set. ChronoBERT is a ModernBERT model and the others are BERT or DeBERTa models; each is read with its own tokenizer.

\begin{table}[tbp]
\centering
\footnotesize
\caption{Readers compared. The released model is shaded.}
\label{tab:readers}
\setlength{\tabcolsep}{4pt}
\begin{tabular}{@{}L{3.5cm}L{5.6cm}L{5.7cm}@{}}
\toprule
Group & Readers & Training text and its dates \\
\midrule
Word overlap & TF-IDF, BM25 & No training. Word weights are fitted on calls up to the cutoff. \\
\addlinespace
Pretrained encoders, zero-shot & NoLBERT, ChronoBERT-2012, ChronoBERT-2023, FinBERT, DeBERTa-v3-base & NoLBERT: text of 1976--1995. ChronoBERT: text up to each year end. FinBERT: pretraining includes calls of 2004--2019. DeBERTa-v3: no date control. \\
\addlinespace
Off-the-shelf retrievers & all-mpnet-base-v2, BGE-base-en-v1.5, E5-base-v2 & Web-scale training data with no date control. \\
\addlinespace
Same recipe, calls up to 2023 & NoLBERT-, ChronoBERT-, FinBERT- and DeBERTa-Calls-2023 & US calls up to 2023; one 150-minute run on one T4 each. \\
\addlinespace
Same recipe, calls up to 2012 & NoLBERT-Calls-2012, ChronoBERT-Calls-2012 & US calls up to 2012; one 150-minute run on one T4 each. \\
\addlinespace
Our staged model & NoLBERT-Calls-2012, stage 1 & Calls from all countries up to 2012; 480 minutes on one RTX 3060. \\
\rowcolor{softteal}
 & \textbf{NoLBERT-Calls-2023 (released)} & Stage 1, then US calls up to 2023 in two 150-minute runs on one T4. \\
 
\bottomrule
\end{tabular}
\end{table}

\subsection{A Point-in-Time Encoder}
\label{sec:methods:nolbert}

\begin{figure}[tbp]
\centering
\resizebox{\linewidth}{!}{\input{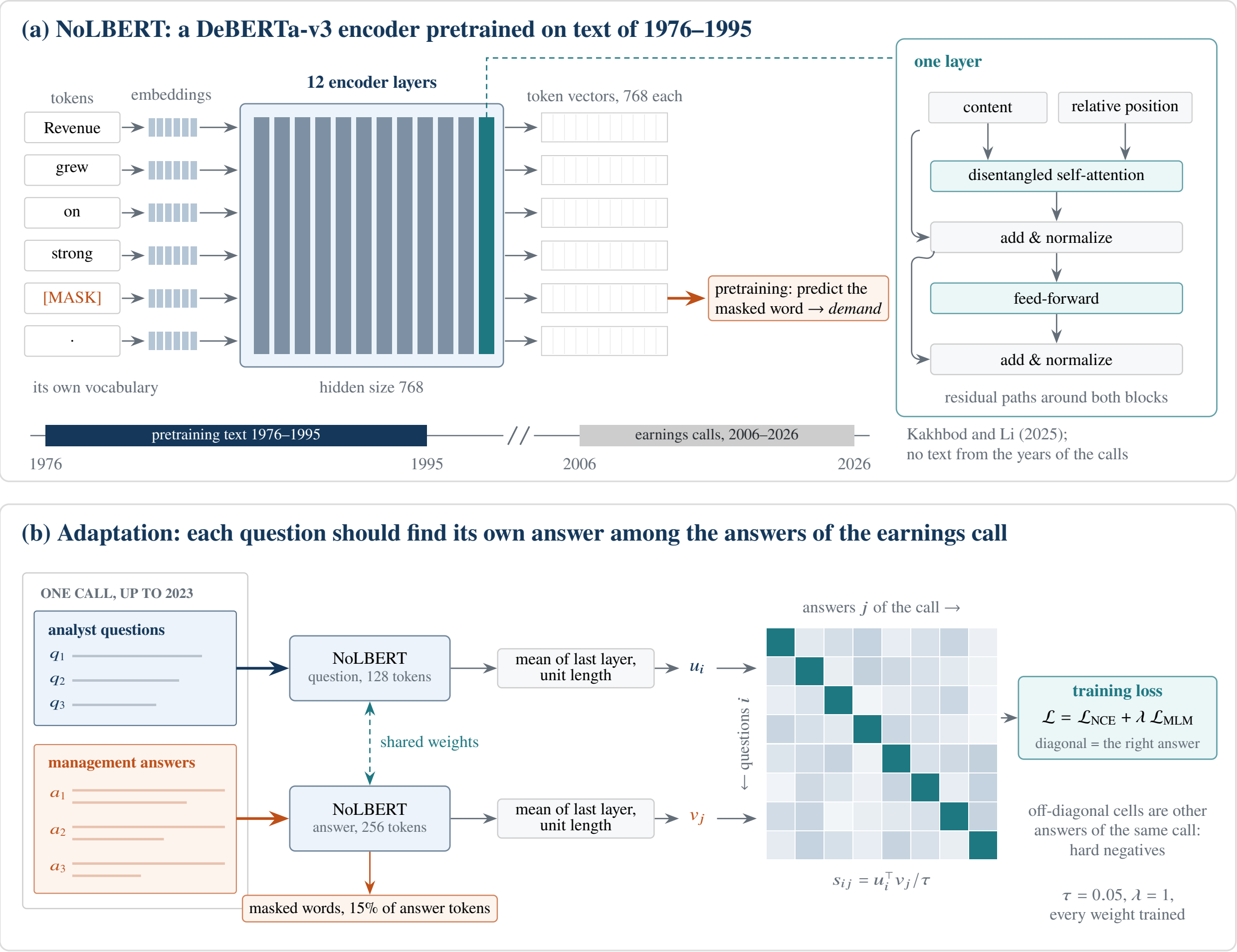}}
\caption{NoLBERT and its adaptation to calls. (a) The pretrained encoder of \citet{kakhbod2025nolbert}, a DeBERTa-v3-base model trained to predict masked words in text of 1976--1995. (b) One training step. The questions and answers of a call pass through the same encoder, and the loss rewards placing each question closest to its own answer. The other answers of the call serve as hard negatives.}
\label{fig:nolbert}
\end{figure}

NoLBERT \citep{kakhbod2025nolbert} is a DeBERTa-v3-base encoder \citep{he2021debertav3} pretrained only on text published between 1976 and 1995 (Figure~\ref{fig:nolbert}a). It has its own vocabulary and its own masked-word head, so no later text enters through either. What the model knows about the years of our calls is therefore exactly what we choose to show it. NoLBERT was trained to predict masked words, however, not to compare two texts, and it has never seen the language of a modern earnings call. Used as it is, it ranks the true answer first for 26.3\% of test questions. The adaptation has to supply both missing skills.

\subsection{Adapting NoLBERT to Calls}
\label{sec:methods:finetune}

Nothing from outside the transcripts enters the adaptation. The calls provide their own training signal: an analyst's question and the answer that follows it form a natural pair, and so do two passages of the same call.

\paragraph{Pairs}
Three quarters of the batches hold question--answer pairs. We pack each of these batches call by call, so that several exchanges of one call sit in the same batch and the model has to tell apart answers given by the same managers in the same quarter. Such negatives are much harder than answers from unrelated calls. The remaining quarter of the batches pair two passages of the same call.

\paragraph{Objective}
For a text $x$ of $T$ tokens, not counting padding, let $h_t(x)$ be the last-layer vector of token $t$. The embedding of $x$ is the mean of these vectors, normalized to unit length:
\begin{equation}
\label{eq:embed}
e(x) \;=\; \frac{\bar h(x)}{\lVert \bar h(x) \rVert},
\qquad
\bar h(x) \;=\; \frac{1}{T} \sum_{t=1}^{T} h_t(x).
\end{equation}
A batch holds $B$ pairs $(q_i, a_i)$, $i = 1, \ldots, B$. With $u_i = e(q_i)$, $v_i = e(a_i)$ and $s_{ij} = u_i^{\top} v_j / \tau$, the symmetric InfoNCE loss \citep{oord2018cpc,gao2021simcse} asks each question to find its answer and each answer to find its question:
\begin{equation}
\label{eq:infonce}
\mathcal{L}_{\mathrm{NCE}}
\;=\; -\frac{1}{2B} \sum_{i=1}^{B}
\left[\, \log \frac{e^{s_{ii}}}{\sum_{j=1}^{B} e^{s_{ij}}}
\;+\; \log \frac{e^{s_{ii}}}{\sum_{j=1}^{B} e^{s_{ji}}} \,\right],
\qquad \tau = 0.05 .
\end{equation}
We add a masked-word loss on the answer or passage side, computed with NoLBERT's own head. With $M$ the set of masked positions (15\% of the tokens) and $\tilde a$ the masked text, the loss is
\begin{equation}
\label{eq:loss}
\mathcal{L} \;=\; \mathcal{L}_{\mathrm{NCE}} + \lambda\, \mathcal{L}_{\mathrm{MLM}},
\qquad
\mathcal{L}_{\mathrm{MLM}} \;=\; -\frac{1}{|M|} \sum_{t \in M} \log p\!\left(a_t \mid \tilde a\right),
\qquad \lambda = 1 .
\end{equation}
The contrastive term teaches the model to compare texts and the masked-word term teaches it the vocabulary of calls.

\paragraph{Stage 1: calls up to 2012}
Starting from NoLBERT, we train on calls from all countries up to 31 December 2012, with up to eight exchanges of a call per batch, batches of 32 and 128 tokens on each side. We use AdamW with a learning rate of $2\times10^{-5}$, 5\% warm-up and cosine decay, on one RTX 3060 for 480 minutes, about 19,000 steps. A checkpoint is scored every 15 minutes on 300 calls from 2013--2015. The best one, at step 17,451, has a validation top-1 of 70.3\% (Figure~\ref{fig:adaptation}a).

\paragraph{Stage 2: US calls up to 2023}
Stage~2 continues from stage~1 on US calls up to 31 December 2023 with the same objective and two changes. Batches hold up to 32 exchanges of the same call, so that far more of the negatives are hard ones, and answers are read at 256 tokens instead of 128, so that the model sees more of each long answer. Stage~2 consists of two runs of 150 minutes each on one T4, the second starting from the first with a fresh optimizer and schedule. Validation top-1 on the 300 calls of 2024 reaches 80.3\% after the first run and 80.7\% after the second, which is the released model. Table~\ref{tab:hyper} lists all settings. The released weights are stored in half precision and give the same embeddings as the full-precision checkpoint.

\paragraph{Model selection}
Every choice is made on the validation calls: the checkpoint, the stage-2 recipe, the blend weight of Section~\ref{sec:methods:blend} and the model we release. During training we score call-weighted top-1 on the validation calls every 15 minutes and keep the best checkpoint. Section~\ref{sec:results:cost} reports every stage-2 run we made.

\subsection{Within-Call Answer Retrieval}
\label{sec:methods:retrieval}

The main task turns the training signal into a test. For each analyst question $q$ of a call $c$, a reader scores every management answer of the call, and we record the rank of the answer that was actually given. With $s(q, a)$ the cosine similarity of the two embeddings, $a_q$ the true answer and $A_c$ the answers of the call, the rank is
\begin{equation}
\label{eq:rank}
r_q \;=\; 1 + \#\bigl\{\, a \in A_c \setminus \{a_q\} \;:\; s(q, a) \ge s(q, a_q) \,\bigr\},
\end{equation}
so a tie counts against the reader. Questions are cut at 128 tokens and answers at 128 tokens, or at 256 tokens for the released model and for the rows marked so. Over the set $Q$ of test questions we report top-1 and the mean reciprocal rank,
\begin{equation}
\label{eq:metrics}
\text{Top-1} \;=\; \frac{1}{|Q|} \sum_{q \in Q} \mathbf{1}\{r_q = 1\},
\qquad
\text{MRR} \;=\; \frac{1}{|Q|} \sum_{q \in Q} \frac{1}{r_q},
\end{equation}
together with top-3, defined like top-1 with $r_q \le 3$, and call-weighted top-1, which averages top-1 within each call first so that every call counts the same. A reader that guesses scores the mean of $1/|A_c|$, which is 8.0\% on the test.

The task has three useful properties. It needs no labels, so it can be run on every call. Its candidates come from one firm and one quarter, so recognizing the firm, which \citet{yu2025dec} find to dominate transcript representations, does not help. And it measures something researchers care about: an answer that a good reader cannot match to its question is a candidate non-answer \citep{gow2021nonanswers}. Since the task mirrors the training objective, the main test measures how well the trained skill carries to later calls; the tasks of Section~\ref{sec:methods:tasks} measure transfer to other uses.

\subsection{Blending with Word Overlap}
\label{sec:methods:blend}

Encoders and word-overlap readers make different mistakes, so we also report a blend of the two. For question $q$ and candidate answer $a$,
\begin{equation}
\label{eq:blend}
s^{\text{blend}}_{qa} \;=\; w\, z\!\left(s^{\text{tfidf}}_{qa}\right) + (1-w)\, z\!\left(s^{\text{model}}_{qa}\right),
\end{equation}
where $z(\cdot)$ standardizes a score over the candidate answers of the question. The weight $w \in \{0, 0.1, \ldots, 1\}$ maximizes call-weighted top-1 on the validation calls; it is 0.1 for the released model. TF-IDF weights are fitted on 20,000 random calls up to the cutoff.

\subsection{Time Test}
\label{sec:methods:time}

The time test asks whether a model gains from having been trained on the years it is tested on. Every reader is scored on 500 random US calls a year from 2013 to 2024 and on the main test for 2025 and 2026. The models adapted through 2023 were trained on calls from 2013--2023 but not on calls from 2025--2026. If training on a year helps a model on that year, the lead of these models over a model with an earlier cutoff should be larger on the first period than on the second.

Let $A_{m,y}$ be the top-1 of model $m$ on the calls of year $y$ and let $m_0$ be NoLBERT adapted through 2012 with the shared recipe. The in-sample premium of $m$ is its average lead over $m_0$ on the years it trained on minus its average lead on the years after its cutoff,
\begin{equation}
\label{eq:premium}
\pi_m \;=\; \frac{1}{11} \sum_{y=2013}^{2023} \bigl(A_{m,y} - A_{m_0,y}\bigr)
\;-\; \frac{1}{2} \sum_{y=2025}^{2026} \bigl(A_{m,y} - A_{m_0,y}\bigr).
\end{equation}
Both models are scored on the same calls in every year, so the comparison holds the calls fixed. Because leads can drift over a decade, we also compute the premium against the three years just before the cutoff, replacing 2013--2023 by 2021--2023 in Equation~(\ref{eq:premium}). Two readers with a 2012 cutoff have read neither period and serve as placebos: their premiums show how far the statistic moves when there is nothing to find. Intervals come from 2,000 paired bootstrap resamples of calls within each year.

\subsection{Other Tasks}
\label{sec:methods:tasks}

Five further tasks test whether adaptation helps beyond the task it was trained for. The first three use the representations as they are, on calls from 2025--2026.

\emph{Passage matching.} A passage of prepared remarks must find an answer passage of its own call among 1,000 calls; chance top-1 is 0.1\%.

\emph{Speaker role.} A logistic probe separates analyst questions from executive answers. It is fitted on calls up to 2023, its penalty is chosen on 2024 and it is tested on 2025--2026.

\emph{Masked finance terms.} We mask 100 single-token finance terms in 3,000 passages and score the model's own masked-word head.

\emph{SubjECTive-QA.} We run the benchmark of \citet{pardawala2024subjectiveqa} in two ways. A logistic probe reads frozen embeddings, and a fresh classification head is trained together with the full encoder (AdamW at $2\times10^{-5}$, three epochs, the checkpoint with the lowest validation loss). Question and answer are read at 256 tokens, on the authors' three splits.

\emph{GLUE.} To check that adaptation does not cost general language ability, we fine-tune on CoLA, QNLI and SST-2 \citep{wang2019glue} with batches of 32 and 128-token inputs, tune on a grouped 10\% split of the training data, and score the official development sets.

\subsection{Inference}
\label{sec:methods:evaluation}

Questions of one call are not independent, so intervals resample calls. Top-1 intervals use 1,000 bootstrap resamples of calls. To compare two readers we pair them on the same questions and take the interval of the difference from 2,000 resamples of calls. With 20,323 test calls, the intervals are narrow enough to detect differences of a few tenths of a point. They reflect the sampling of test calls, not randomness in training: each run uses one seed.

\section{Results}
\label{sec:results}

\subsection{Answer Retrieval on 2025--2026 Calls}
\label{sec:results:retrieval}

Table~\ref{tab:retrieval} reports the main test. NoLBERT-Calls-2023 ranks the true answer first for 78.0\% of the 254,575 test questions (95\% interval 77.8--78.2) and among the top three for 94.0\%, with a mean reciprocal rank of 86.4. Chance is 8.0\%. Blended with word overlap, the model reaches 79.0\%. All candidates in a ranking come from one call, and the model still picks the right answer more than three times in four.

The closest competitor is ChronoBERT adapted with the shared recipe and read with 256-token answers, at 77.0\%; the released model leads it by 1.0 point (0.9--1.2, paired). It leads FinBERT-Calls-2023 by 3.4 points (3.2--3.5), or by 3.9 if FinBERT-Calls also reads 256-token answers. Retrievers and word overlap are far behind: E5, the best retriever, by 18.7 points and BM25 by 19.8. Blending with word overlap on both sides changes little. The released model's lead is then 1.1 points (0.9--1.2) over ChronoBERT-Calls at 256 tokens and 1.9 points (1.7--2.0) over FinBERT-Calls, and the released model without any blend matches or beats every comparison reader with one; the closest, ChronoBERT-Calls at 256 tokens with word overlap, scores 77.9\%.

\begin{table}[tbp]
\centering
\footnotesize
\caption{Within-call answer retrieval on every US call from January 2025 to October 2026 (20,323 calls, 254,575 questions).}
\label{tab:retrieval}
\setlength{\tabcolsep}{4.2pt}
\begin{tabular}{@{}L{5.9cm}lcccl@{}}
\toprule
Reader & \makecell[l]{Top-1\\{\scriptsize[95\% interval]}} & \makecell[c]{+ word\\overlap} & Top-3 & MRR & \makecell[l]{$\Delta$ to released\\{\scriptsize[95\% interval]}} \\
\midrule
\multicolumn{6}{@{}l}{\textit{Adapted on calls up to 2023}} \\
\rowcolor{softteal}
\quad \textbf{NoLBERT-Calls-2023 (released)} & \topA{78.0} {\scriptsize[77.8, 78.2]} & \topA{79.0} & \topA{94.0} & \topA{86.4} & \textemdash \\
\quad ChronoBERT-Calls-2023 at 256 tokens & \topB{77.0} {\scriptsize[76.8, 77.2]} & \topB{77.9} & \topB{93.5} & \topB{85.7} & 1.0 {\scriptsize[0.9, 1.2]} \\
\quad FinBERT-Calls-2023 & \topC{74.7} {\scriptsize[74.5, 74.9]} & \topC{77.1} & \topC{92.8} & \topC{84.2} & 3.4 {\scriptsize[3.2, 3.5]} \\
\quad ChronoBERT-Calls-2023 & 72.3 {\scriptsize[72.1, 72.6]} & 74.6 & 91.2 & 82.4 & 5.7 {\scriptsize[5.5, 5.9]} \\
\quad DeBERTa-Calls-2023 & 68.2 {\scriptsize[68.0, 68.4]} & 74.4 & 89.6 & 79.7 & 9.8 {\scriptsize[9.6, 10.0]} \\
\quad NoLBERT-Calls-2023, same recipe & 67.0 {\scriptsize[66.7, 67.2]} & 70.6 & 88.1 & 78.6 & 11.1 {\scriptsize[10.9, 11.2]} \\
\addlinespace[2pt]
\multicolumn{6}{@{}l}{\textit{Adapted on calls up to 2012}} \\
\quad NoLBERT-Calls-2012, stage 1 & 71.4 {\scriptsize[71.2, 71.6]} & 73.9 & 90.6 & 81.7 & 6.7 {\scriptsize[6.5, 6.8]} \\
\quad ChronoBERT-Calls-2012 & 69.3 {\scriptsize[69.1, 69.5]} & 72.5 & 89.5 & 80.3 & 8.8 {\scriptsize[8.6, 8.9]} \\
\quad NoLBERT-Calls-2012, same recipe & 65.1 {\scriptsize[64.9, 65.3]} & 69.5 & 86.8 & 77.1 & 12.9 {\scriptsize[12.8, 13.1]} \\
\addlinespace[2pt]
\multicolumn{6}{@{}l}{\textit{Off-the-shelf retrievers}} \\
\quad E5-base-v2 & 59.3 {\scriptsize[59.1, 59.5]} & 65.3 & 82.4 & 72.5 & 18.7 {\scriptsize[18.5, 18.9]} \\
\quad BGE-base-en-v1.5 & 58.1 {\scriptsize[57.9, 58.3]} & 64.0 & 80.8 & 71.3 & 19.9 {\scriptsize[19.7, 20.2]} \\
\quad all-mpnet-base-v2 & 57.5 {\scriptsize[57.2, 57.7]} & 64.8 & 81.4 & 71.2 & 20.6 {\scriptsize[20.4, 20.8]} \\
\addlinespace[2pt]
\multicolumn{6}{@{}l}{\textit{Word overlap}} \\
\quad BM25 & 58.2 {\scriptsize[58.0, 58.5]} & \textemdash & 80.7 & 71.2 & 19.8 {\scriptsize[19.6, 20.0]} \\
\quad TF-IDF & 57.3 {\scriptsize[57.0, 57.5]} & \textemdash & 80.4 & 70.6 & 20.8 {\scriptsize[20.6, 21.0]} \\
\addlinespace[2pt]
\multicolumn{6}{@{}l}{\textit{Pretrained encoders, zero-shot}} \\
\quad FinBERT & 31.9 {\scriptsize[31.7, 32.1]} & 59.4 & 59.1 & 49.9 & 46.2 {\scriptsize[45.9, 46.4]} \\
\quad NoLBERT & 26.3 {\scriptsize[26.1, 26.5]} & 58.6 & 51.7 & 44.4 & 51.7 {\scriptsize[51.5, 51.9]} \\
\quad ChronoBERT-2023 & 24.1 {\scriptsize[23.9, 24.2]} & 58.2 & 48.1 & 41.8 & 54.0 {\scriptsize[53.7, 54.2]} \\
\quad ChronoBERT-2012 & 23.7 {\scriptsize[23.6, 23.9]} & 58.2 & 47.3 & 41.3 & 54.3 {\scriptsize[54.1, 54.5]} \\
\quad DeBERTa-v3-base & 10.0 {\scriptsize[9.9, 10.1]} & 57.3 & 27.9 & 26.9 & 68.0 {\scriptsize[67.8, 68.2]} \\
\addlinespace[2pt]
\quad Chance & 8.0 & & & & \\
 
\bottomrule
\end{tabular}
\tablenotes{Top-1, top-3 and MRR in percent, weighting every question equally; intervals from 1,000 bootstrap resamples of calls. \emph{+ word overlap}: top-1 of the reader blended with TF-IDF, with the weight chosen on the 2024 validation calls. $\Delta$: the released model's top-1 minus the row's, in points, on the same questions, with a paired interval from 2,000 resamples of calls. \emph{Same recipe}: one 150-minute run on one T4 with shared settings. Table~\ref{tab:retrieval-full} adds call-weighted top-1 and further rows. \shadekey.}
\end{table}

\begin{figure}[tbp]
\centering
\includegraphics[width=\linewidth]{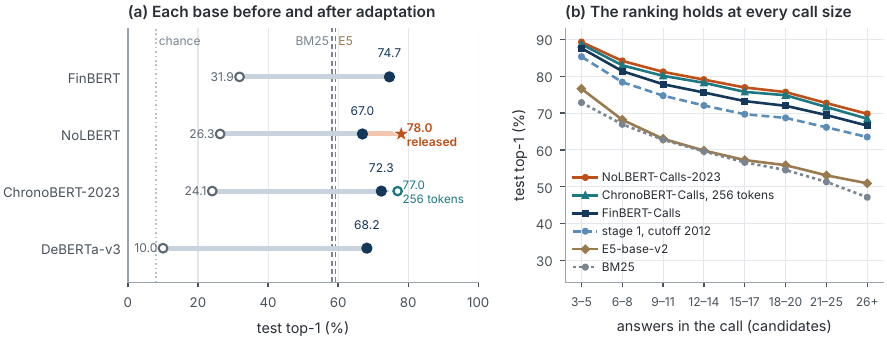}
\caption{What drives accuracy on the 2025--2026 test. (a) Top-1 of each pretrained encoder used zero-shot (open circles) and after adaptation on US calls up to 2023 with the shared 150-minute recipe (filled). The open teal circle reads ChronoBERT-Calls-2023 with 256-token answers, without retraining, and the star is the released model. Dashed lines mark BM25 and E5, the dotted line chance. (b) Top-1 by the number of candidate answers in the call. The released model leads in every bin.}
\label{fig:results}
\end{figure}

Figure~\ref{fig:results}b splits the test by the number of answers in a call. Accuracy falls as calls get larger, from 89.3\% for calls with three to five answers to 69.8\% for calls with more than 25, but the order of the readers never changes. The released model leads ChronoBERT-Calls at 256 tokens by 0.6 to 1.3 points in every size bin. The two test years agree, at 77.8\% for 2025 and 78.3\% for 2026, and ranking the readers by call-weighted top-1 gives exactly the same order as ranking them by question-weighted top-1 (Tables~\ref{tab:time} and~\ref{tab:retrieval-full}).

\subsection{What Drives Accuracy}
\label{sec:results:drivers}

Used as they are, encoders read calls poorly (Figure~\ref{fig:results}a). Every pretrained encoder scores below 32\%, far under the 58.2\% of BM25, and a more recent pretraining corpus does not help. ChronoBERT-2023, whose text runs to 2023, scores 24.1\%, below NoLBERT at 26.3\% and level with ChronoBERT-2012 at 23.7\%. FinBERT, whose pretraining includes calls of 2004--2019, reaches only 31.9\%. Adaptation changes the picture. With the shared recipe and 150 minutes of training, the four encoders gain between 41 and 58 points and end between 67.0\% and 74.7\%, well above every retriever.

The length of the answer window is nearly as important. Answers in recent calls are long (Section~\ref{sec:data:calls}), and 128 tokens cover only the start of many of them. Reading the same ChronoBERT-Calls-2023 checkpoint with 256-token answers, without any retraining, lifts it from 72.3\% to 77.0\%. FinBERT-Calls does not gain from the longer window (74.1\%). Our stage-1 model gains 4.5 points of validation top-1 from it before any further training, from 73.8\% to 78.3\% (Figure~\ref{fig:adaptation}b).

Word overlap adds little to a strong reader. The blend lifts the zero-shot encoders by 27 to 47 points, the retrievers by 6 to 7 and the same-recipe encoders by 2 to 6, but the released model by only 1.0 point. Most of what shared words can tell this model, it has already learned.

\subsection{The Cost of the Cutoff}
\label{sec:results:cost}

\begin{figure}[tbp]
\centering
\includegraphics[width=\linewidth]{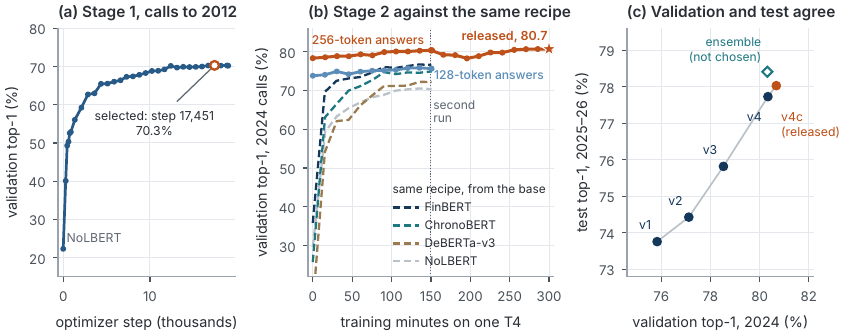}
\caption{How the released model was built. (a) Stage~1: validation top-1 on 300 calls from 2013--2015, pooled over questions, during the 480-minute run; the circle marks the selected checkpoint. (b) Validation top-1 on the 300 calls of 2024, weighting calls equally, against training minutes on one T4. Dashed lines: each base adapted with the shared recipe. Solid lines: stage~2, starting from stage~1, with 128-token answers (run v1) and with 256-token answers (run v4, then a second run, which is the released model). (c) Validation and test top-1 of the five stage-2 runs. An ensemble of v4 and v3 scored higher on the test but lower on validation and was not chosen.}
\label{fig:adaptation}
\end{figure}

The shared recipe makes one result plain: NoLBERT is not the best starting point. Adapted for 150 minutes, it reaches 67.0\%, behind DeBERTa-v3 at 68.2\%, ChronoBERT at 72.3\% and FinBERT at 74.7\%. The released model's lead therefore comes from somewhere else, namely more training and a recipe developed for it on the validation calls (Figure~\ref{fig:adaptation} and Table~\ref{tab:development}).

Stage~1 ran for 480 minutes on calls from all countries up to 2012. On its own it scores 71.4\% on the test, above both same-recipe models with the same cutoff (69.3\% for ChronoBERT-Calls-2012 and 65.1\% for NoLBERT-Calls-2012) and above two of the four same-recipe models adapted through 2023. Stage~2 then starts from stage~1 rather than from the pretrained base, and Figure~\ref{fig:adaptation}b shows what that head start is worth. At minute zero, stage~1 already scores 73.8\% on the 2024 validation calls with 128-token answers, close to where the best same-recipe runs end after 150 minutes, and 78.3\% with 256-token answers, more than any of them reaches.

We made five stage-2 runs, each of 150 minutes. Packing up to 32 exchanges of a call into one batch, so that most negatives come from the same call, added 1.3 points of validation top-1 (run v2). On top of that, batches of question--answer pairs only with a higher learning rate added 1.4 points (v3), and reading answers at 256 tokens added 3.2 (v4). We continued the best run, v4, for a second 150 minutes, which added 0.3 points more. Its best checkpoint is its last, so more training would probably help. Validation and test agree throughout (Figure~\ref{fig:adaptation}c): the 2024 calls rank the five runs exactly as the test does, with validation 2.0 to 2.7 points higher. The one disagreement is an ensemble of runs v4 and v3, which is lower on validation (80.3\% against 80.7\%) but higher on the test (78.4\% against 78.0\%). We did not release it, because choosing it would have meant choosing on the test.

In total the released model took about 13 GPU-hours, five times the 150 minutes of a same-recipe run. This is the cost of the cutoff: a base that has read nothing after 1995 needs more adaptation to overtake encoders pretrained on recent text, but it gets there. We did not try the tuned recipe on the other encoders, and with it they might close part of the gap.

\subsection{No Look-Ahead Premium}
\label{sec:results:time}

\begin{figure}[tbp]
\centering
\includegraphics[width=\linewidth]{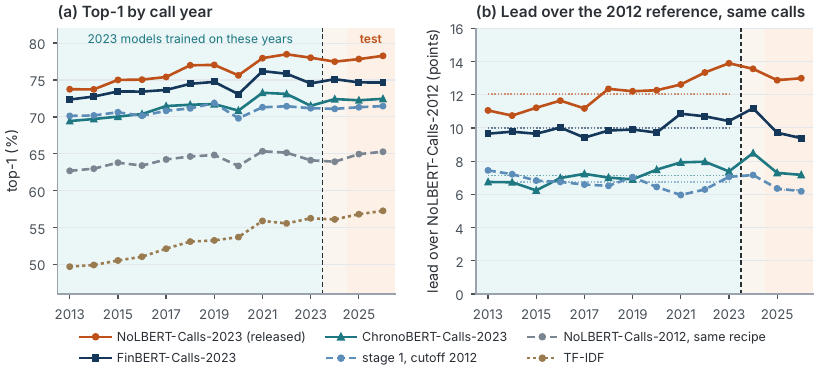}
\caption{The time test. (a) Top-1 by call year, on 500 US calls a year for 2013--2024 and on the full main test for 2025 and 2026. Teal shading marks the years the 2023 models were trained on, light orange 2024 (validation) and orange the test years. (b) Each reader's lead over NoLBERT-Calls-2012 (same recipe) on the same calls. Thin dotted lines show the average lead over 2013--2023 and over 2025--2026. The stage-1 model has a 2012 cutoff and has read neither period.}
\label{fig:time}
\end{figure}

Figure~\ref{fig:time} and Table~\ref{tab:premium} report the time test. Every reader's accuracy drifts up with the year of the call, most of all for TF-IDF (from 49.7\% in 2013 to 57.3\% in 2026), because later calls have fewer exchanges and so fewer candidates: chance top-1 rises from 5.6\% in 2013 to 8.0\% in 2025--2026 (Table~\ref{tab:time}). Comparing each reader with a reference on the same calls removes this drift.

Models with a 2012 cutoff do not decay. Stage~1 scores 70.1\% on calls from 2013 and 71.5\% on calls from 2026, fourteen years after its cutoff, although the subjects of calls changed a great deal in between (Figure~\ref{fig:data}b). They do fall behind models trained on later calls. The released model's lead over stage~1 widens from 3.6 points in 2013 to 6.8 points in 2023. Part of this may reflect the longer answers of recent calls, which the released model reads at 256 tokens and stage~1 at 128.

The look-ahead question is whether a model's lead is larger on the years it trained on. Against NoLBERT-Calls-2012, the released model leads by 12.0 points on 2013--2023 and by 12.9 points on 2025--2026, a premium of $-0.9$ points ($-1.2$ to $-0.5$): it is relatively stronger on calls it never saw. Because leads grow over the training years, an average over 2013--2023 could hide a premium concentrated in the last few of them. Against 2021--2023 alone, the released model's premium is $+0.3$ points ($-0.3$ to $+1.0$), close to the $+0.2$ of the two placebo models that read neither period. FinBERT-Calls and DeBERTa-Calls show the largest premiums by this measure, $+1.1$ (0.5--1.8) and $+1.0$ (0.2--1.8) points. On this task, then, look-ahead is worth at most about a point to any encoder we adapted, small next to the 41 to 58 points that adaptation adds, and for the released model it cannot be told apart from zero.

\begin{table}[tbp]
\centering
\footnotesize
\caption{Look-ahead premium in the time test.}
\label{tab:premium}
\setlength{\tabcolsep}{4.2pt}
\begin{tabular}{@{}L{5.1cm}rrrll@{}}
\toprule
 & \multicolumn{3}{c}{Lead over NoLBERT-Calls-2012} & \multicolumn{2}{c}{Premium {\scriptsize[95\% interval]}} \\
\cmidrule(lr){2-4}\cmidrule(l){5-6}
Reader & 2013--23 & 2021--23 & 2025--26 & 2013--23 vs 2025--26 & 2021--23 vs 2025--26 \\
\midrule
\multicolumn{6}{@{}l}{\textit{Adapted on calls up to 2023}} \\
\rowcolor{softteal}
\quad \textbf{NoLBERT-Calls-2023 (released)} & +12.0 & +13.3 & +12.9 & $-$0.9 {\scriptsize[$-$1.2, $-$0.5]} & +0.3 {\scriptsize[$-$0.3, +1.0]} \\
\quad FinBERT-Calls-2023 & +10.0 & +10.7 & +9.6 & +0.4 {\scriptsize[+0.1, +0.8]} & +1.1 {\scriptsize[+0.5, +1.8]} \\
\quad ChronoBERT-Calls-2023 & +7.1 & +7.8 & +7.2 & $-$0.1 {\scriptsize[$-$0.4, +0.3]} & +0.5 {\scriptsize[$-$0.1, +1.1]} \\
\quad DeBERTa-Calls-2023 & +3.1 & +4.1 & +3.1 & +0.0 {\scriptsize[$-$0.4, +0.5]} & +1.0 {\scriptsize[+0.2, +1.8]} \\
\quad NoLBERT-Calls-2023, same recipe & +1.3 & +2.1 & +1.8 & $-$0.6 {\scriptsize[$-$0.8, $-$0.3]} & +0.3 {\scriptsize[$-$0.2, +0.8]} \\
\addlinespace[2pt]
\multicolumn{6}{@{}l}{\textit{Placebo: adapted on calls up to 2012}} \\
\quad NoLBERT-Calls-2012, stage 1 & +6.7 & +6.4 & +6.3 & +0.5 {\scriptsize[+0.2, +0.8]} & +0.2 {\scriptsize[$-$0.4, +0.7]} \\
\quad ChronoBERT-Calls-2012 & +4.8 & +4.4 & +4.2 & +0.6 {\scriptsize[+0.3, +1.0]} & +0.2 {\scriptsize[$-$0.4, +0.9]} \\
\addlinespace[2pt]
\multicolumn{6}{@{}l}{\textit{Pretraining includes calls of 2004--2019}} \\
\quad FinBERT, zero-shot & $-$33.8 & $-$33.6 & $-$33.3 & $-$0.6 {\scriptsize[$-$1.0, $-$0.1]} & $-$0.3 {\scriptsize[$-$1.1, +0.5]} \\
 
\bottomrule
\end{tabular}
\tablenotes{Leads are differences in top-1, in points, between the reader and NoLBERT-Calls-2012 (same recipe) on the same calls, averaged over the years shown. The premium is the average lead over the in-sample years minus the average lead over 2025--2026; a positive premium means the reader does better on the years it was trained on. Intervals from 2,000 paired bootstrap resamples of calls within each year. Placebo readers have a 2012 cutoff and have read neither period.}
\end{table}

\subsection{Beyond Answer Retrieval}
\label{sec:results:tasks}

On tasks other than the one it was trained for, the released model is competitive but rarely first (Tables~\ref{tab:tasks} and \ref{tab:finetuned}). Passage matching is hard for every encoder. Adaptation lifts NoLBERT from 0.8\% to 7.6\% top-1 (chance is 0.1\%), but TF-IDF is best at top-1 with 12.6\% and FinBERT-Calls at top-10 with 27.9\%. On masked finance terms, adaptation raises NoLBERT from 18.7\% to 25.6\% top-1, level with ChronoBERT-Calls, while the FinBERT models lead at 38--40\%, helped by a vocabulary built from financial text. Speaker role and the SubjECTive-QA probe barely separate the readers: balanced accuracy on role lies between 88\% and 98\% for all of them, and the probe scores between 63 and 67, against 55.6 for the majority class, with TF-IDF as good as any encoder.

\begin{table}[tbp]
\centering
\footnotesize
\caption{Tasks on frozen representations, calls of 2025--2026 unless noted.}
\label{tab:tasks}
\setlength{\tabcolsep}{4pt}
\begin{tabular}{@{}L{5.0cm}rrrrrr@{}}
\toprule
 & \multicolumn{2}{c}{Passage matching} & & \multicolumn{2}{c}{Masked terms} & \\
\cmidrule(lr){2-3}\cmidrule(lr){5-6}
Reader & Top-1 & Top-10 & Role & Top-1 & Top-5 & \makecell[c]{SubjECTive-QA\\probe} \\
\midrule
\rowcolor{softteal}
\textbf{NoLBERT-Calls-2023 (released)} & \topC{7.6} & \topC{21.9} & 96.2 & \topC{25.6} & \topC{47.1} & 64.8 \\
NoLBERT-Calls-2012, stage 1 & 5.8 & 17.9 & 95.9 & 23.9 & 45.1 & 65.4 \\
NoLBERT-Calls-2023, same recipe & 6.3 & 15.8 & 95.6 & 21.3 & 40.2 & 65.1 \\
FinBERT-Calls-2023 & \topB{8.2} & \topA{27.9} & \topB{97.8} & \topA{39.6} & \topA{65.0} & \topA{66.5} \\
ChronoBERT-Calls-2023 & 6.6 & 19.2 & 95.4 & 25.2 & \topC{47.1} & 65.4 \\
DeBERTa-Calls-2023 & 2.7 & 15.2 & \topC{97.2} & 20.4 & 41.2 & 65.0 \\
E5-base-v2 & 5.2 & 14.0 & 96.5 & \textemdash & \textemdash & \topC{66.2} \\
TF-IDF & \topA{12.6} & \topB{26.0} & 88.1 & \textemdash & \textemdash & \topB{66.3} \\
NoLBERT, zero-shot & 0.8 & 4.1 & 96.6 & 18.7 & 34.6 & 65.3 \\
FinBERT, zero-shot & 0.9 & 6.9 & \topA{98.2} & \topB{38.0} & \topB{62.6} & 66.0 \\
ChronoBERT-2023, zero-shot & 0.5 & 3.1 & 96.1 & 23.6 & 43.5 & 64.8 \\
\addlinespace[2pt]
\textit{Chance or majority class} & 0.1 & 1.0 & 50.0 & & & 55.6 \\
 
\bottomrule
\end{tabular}
\tablenotes{All figures in percent. \emph{Passage matching}: a prepared-remarks passage must find an answer passage of its own call among 1,000 calls. \emph{Role}: balanced accuracy of a logistic probe separating analyst questions from executive answers. \emph{Masked terms}: macro-averaged recovery of 100 single-token finance terms in 3,000 passages, which needs a masked-word head. \emph{SubjECTive-QA probe}: weighted F1 of a logistic probe on frozen embeddings, averaged over six properties and the authors' three splits. Table~\ref{tab:tasks-full} lists every reader. \shadekey.}
\end{table}

Fine-tuning the whole encoder tells a similar story. On SubjECTive-QA, adaptation adds 1.3 to 1.4 points of weighted F1 over base NoLBERT, from 64.4 to 65.7 and 65.8, mostly on the properties Optimistic ($+4.0$) and Specific ($+1.5$). The adapted encoders span 65.5 to 66.2, less than the standard deviation across the three splits, so we do not rank them. On Clear and Relevant almost every head predicts the majority class (Table~\ref{tab:sqa-props}), and on the four properties that carry signal the released model averages 56.8 against 54.8 for base NoLBERT and 57.4 for FinBERT-Calls. On GLUE, adaptation leaves QNLI and SST-2 at or slightly above base NoLBERT and lifts CoLA from 14.8 to 31.9, so reading calls did not cost general language ability. ChronoBERT-2023, pretrained on far more and far more recent text, is ahead on all three tasks, a gap that belongs to the base model.

\begin{table}[tbp]
\centering
\footnotesize
\caption{Benchmarks after fine-tuning the whole encoder.}
\label{tab:finetuned}
\setlength{\tabcolsep}{5pt}
\begin{tabular}{@{}L{5.4cm}ccrrr@{}}
\toprule
 & \multicolumn{2}{c}{SubjECTive-QA, weighted F1} & \multicolumn{3}{c}{GLUE, development sets} \\
\cmidrule(lr){2-3}\cmidrule(l){4-6}
Encoder & Six properties (sd) & Four properties & CoLA & QNLI & SST-2 \\
\midrule
NoLBERT, zero-shot base & 64.4 (0.9) & 54.8 & 14.8 & 88.0 & 85.9 \\
NoLBERT-Calls-2023, same recipe & 65.7 (1.0) & 56.8 & 26.8 & 88.2 & 86.4 \\
\rowcolor{softteal}
\textbf{NoLBERT-Calls-2023 (released)} & 65.8 (1.5) & 56.8 & 31.9 & 88.2 & 86.8 \\
FinBERT-Calls-2023 & 66.2 (0.9) & 57.4 & \textemdash & \textemdash & \textemdash \\
ChronoBERT-Calls-2023 & 65.6 (1.0) & 56.5 & \textemdash & \textemdash & \textemdash \\
DeBERTa-Calls-2023 & 65.5 (1.5) & 56.4 & \textemdash & \textemdash & \textemdash \\
ChronoBERT-2023, base & \textemdash & \textemdash & 45.9 & 91.3 & 92.0 \\
 
\bottomrule
\end{tabular}
\tablenotes{\emph{SubjECTive-QA}: a fresh classification head trained with the encoder, mean over the authors' three splits with the standard deviation across splits in brackets. \emph{Four properties}: Assertive, Cautious, Optimistic and Specific; on Clear and Relevant almost every head predicts the majority class. \emph{GLUE}: Matthews correlation for CoLA and accuracy for QNLI and SST-2, times 100, one seed. A dash means not run.}
\end{table}

\subsection{Robustness}
\label{sec:results:robust}

The main ranking does not depend on how the test is weighted or cut. Weighting calls instead of questions leaves the order of all readers unchanged. Splitting the test by call size leaves the order of the readers in Figure~\ref{fig:results}b unchanged, and splitting it by year leaves the order of the readers in Table~\ref{tab:time} unchanged; no reader loses more than 0.2 points from 2025 to 2026. The half-precision release gives the same embeddings as the full-precision checkpoint. One check is missing: each run used a single training seed, so our intervals cover the sampling of test calls but not the randomness of training.

\section{Discussion and Limitations}
\label{sec:discussion}

\paragraph{What a call reader without hindsight costs}
The full adaptation takes about 13 GPU-hours on a consumer GPU and a free-tier T4 (Table~\ref{tab:compute}). Its training signal, the pairing of each analyst question with its answer, is in every transcript, so no labels have to be collected. For that cost a researcher gets a reader that has seen no text after 31 December 2023, that can be used in backtests from 2024 on, and that is more accurate on later calls than every encoder we compared it with, including encoders pretrained on far more recent text. The cost is low enough to repeat the adaptation for other cutoffs. A family of vintages, one for each year, would serve any sample period, in the way ChronoBERT and DatedGPT do for general text \citep{he2025chronologically,yan2026datedgpt}. Our stage-1 model is already one such vintage, with a cutoff at the end of 2012.

\paragraph{What the time test shows and what it does not}
The time test speaks to a narrower question than look-ahead bias in general. It asks whether an encoder adapted on calls from a period reads those calls better than later ones, and for the released model the answer is no. It does not ask whether the encoder knows how later events turned out, which is the main concern with large generative models \citep{lopezlira2025memorization,levy2024caution}. That question is settled by construction: about anything after 31 December 2023, NoLBERT-Calls cannot know more than a reader of that date, because it has read nothing written later. The test also has limits of its own. It uses samples of 500 calls a year, and the leads of all models trend upward over the training years, which is why we report the premium against both the whole training period and its last three years.

\paragraph{Limitations}
Several limits apply to our results. Each run used one training seed, so our intervals cover the sampling of test calls but not training randomness. The stage-2 recipe was tuned for NoLBERT alone, while the other encoders were adapted once with the shared recipe; ChronoBERT, for instance, gains 4.7 points just from reading longer answers and might gain more if retrained with them. Stage~2 and every test use US calls from one provider, so calls from other countries, which stage~1 already reads, and transcripts from other providers are natural next test sets. Answer retrieval measures reading. Whether the gains carry over to economic outcomes, such as volatility or returns after the call, is not tested here. Finally, the license of the transcripts keeps their text out of hosted language-model services, so our comparison covers models that run locally; open-weight generative models that can run under the same terms would be a useful addition.

\paragraph{Uses}
Matching questions to answers gives a direct measure of how well an answer responds to its question, a candidate measure of evasiveness to set beside the non-answer measures of \citet{gow2021nonanswers} and \citet{barth2023nonanswers}. The embeddings can be grouped into topics of discussion, giving each firm and quarter a set of topic shares that can serve as characteristics in asset pricing, and they can be used to build firm-level exposures to new shocks in the spirit of \citet{hassan2019firmlevel}. Each of these measures inherits the cutoff of the model, so each can be backtested without hindsight. These text-derived characteristics could also be incorporated into interpretable machine-learning frameworks such as \citep{bell2026glass,bell2026alphaglass} to predict return and form interpretable portfolios. 

\section{Conclusion}
\label{sec:conclusion}

We present NoLBERT-Calls, an encoder for earnings calls that has read no text dated after 31 December 2023. It starts from NoLBERT, pretrained on text from 1976 to 1995, and is adapted in two stages using only the questions and answers of the calls themselves. On every US call from January 2025 to October 2026, it ranks the true answer to an analyst's question first 78.0\% of the time, against 26.3\% before adaptation, and 79.0\% of the time when blended with word overlap, ahead of every reader we compared. A year-by-year test from 2013 to 2026 finds no look-ahead premium. A point-in-time encoder can thus be adapted to a specialized domain without crossing its cutoff and without giving up accuracy, for about 13 GPU-hours on modest hardware. Research on earnings calls can use it to build measures that are free of hindsight by construction.

\section*{Data and Code Availability}
\label{sec:availability}

The transcripts come from S\&P Capital IQ through WRDS \citep{ciqtranscripts}. They are licensed, and neither the transcripts nor any text derived from them are redistributed; the repository contains no transcript text. The code, aggregate results and the released model, NoLBERT-Calls-2023 in half precision with checksums, are at \url{https://github.com/Shivesh777/NoLBERT-Calls}. The code is under the MIT license and the weights are for research use. SubjECTive-QA (CC BY 4.0) and GLUE are public, and the base model is NoLBERT \citep{kakhbod2025nolbert}.

\clearpage
\bibliographystyle{plainnat}
\bibliography{references}

\clearpage
\appendix
\setcounter{table}{0}
\renewcommand{\thetable}{A\arabic{table}}
\setcounter{figure}{0}
\renewcommand{\thefigure}{A\arabic{figure}}

\section{Additional Results}
\label{app:results}

This appendix gives the full results behind Section~\ref{sec:results}. Table~\ref{tab:time} reports the time test year by year, with the number of questions per call and the chance level in each year. Table~\ref{tab:retrieval-full} lists every reader of the main test, with and without word overlap, including the stage-2 development runs and the ensemble that was not released. Table~\ref{tab:development} describes the stage-2 runs, Table~\ref{tab:tasks-full} gives the tasks on frozen representations for every reader, and Table~\ref{tab:sqa-props} the SubjECTive-QA results property by property.

\begin{table}[!htbp]
\centering
\scriptsize
\caption{The time test by call year: top-1 in percent.}
\label{tab:time}
\setlength{\tabcolsep}{2.6pt}
\begin{tabular}{@{}L{4.7cm}*{14}{r}@{}}
\toprule
Reader & 2013 & 2014 & 2015 & 2016 & 2017 & 2018 & 2019 & 2020 & 2021 & 2022 & 2023 & 2024 & 2025 & 2026 \\
\midrule
TF-IDF (word weights from calls up to 2012) & 49.7 & 49.9 & 50.5 & 51.1 & 52.1 & 53.1 & 53.3 & 53.7 & 55.9 & 55.6 & 56.3 & 56.1 & 56.8 & 57.3 \\
NoLBERT, zero-shot & 22.5 & 22.4 & 22.9 & 23.1 & 24.8 & 24.2 & 25.0 & 24.8 & 25.1 & 25.9 & 25.7 & 25.9 & 26.3 & 26.4 \\
ChronoBERT-2012, zero-shot & 20.7 & 20.5 & 21.6 & 21.6 & 22.9 & 21.8 & 22.1 & 22.4 & 22.3 & 22.5 & 23.2 & 23.7 & 23.8 & 23.6 \\
FinBERT, zero-shot & 29.5 & 28.8 & 28.8 & 30.2 & 30.1 & 29.8 & 30.3 & 30.8 & 31.0 & 31.3 & 31.6 & 31.6 & 31.9 & 31.8 \\
NoLBERT-Calls-2012, stage 1 & 70.1 & 70.2 & 70.6 & 70.2 & 70.8 & 71.2 & 71.9 & 69.8 & 71.3 & 71.4 & 71.2 & 71.1 & 71.3 & 71.5 \\
NoLBERT-Calls-2012, 150-minute stage 1 & 57.6 & 56.9 & 58.5 & 58.5 & 58.5 & 59.7 & 59.2 & 58.1 & 60.3 & 60.5 & 59.9 & 60.0 & 60.5 & 60.7 \\
NoLBERT-Calls-2012, same recipe & 62.7 & 63.0 & 63.8 & 63.4 & 64.2 & 64.6 & 64.8 & 63.4 & 65.3 & 65.1 & 64.1 & 63.9 & 65.0 & 65.3 \\
ChronoBERT-Calls-2012 & 67.7 & 68.4 & 68.7 & 68.6 & 68.7 & 69.3 & 69.6 & 68.4 & 69.4 & 70.2 & 68.2 & 69.0 & 69.2 & 69.4 \\
NoLBERT-Calls-2023, same recipe & 63.1 & 63.5 & 64.5 & 64.3 & 65.0 & 65.8 & 66.2 & 65.4 & 67.4 & 67.1 & 66.5 & 66.6 & 66.8 & 67.1 \\
ChronoBERT-Calls-2023 & 69.4 & 69.7 & 70.0 & 70.4 & 71.5 & 71.6 & 71.7 & 70.8 & 73.3 & 73.1 & 71.5 & 72.4 & 72.3 & 72.5 \\
FinBERT-Calls-2023 & 72.3 & 72.8 & 73.4 & 73.4 & 73.7 & 74.5 & 74.7 & 73.1 & 76.2 & 75.8 & 74.5 & 75.1 & 74.7 & 74.7 \\
DeBERTa-Calls-2023 & 65.1 & 65.1 & 66.5 & 66.2 & 66.6 & 67.3 & 67.5 & 67.5 & 69.4 & 69.7 & 67.8 & 68.0 & 68.3 & 68.1 \\
\rowcolor{softteal}
\textbf{NoLBERT-Calls-2023 (released)} & 73.7 & 73.7 & 75.0 & 75.0 & 75.4 & 77.0 & 77.0 & 75.6 & 78.0 & 78.5 & 78.0 & 77.5 & 77.8 & 78.3 \\
\addlinespace[2pt]
\textit{Questions per call} & 17.7 & 17.3 & 17.2 & 17.0 & 15.8 & 15.4 & 14.9 & 13.8 & 13.6 & 12.8 & 12.9 & 12.5 & 12.6 & 12.5 \\
\textit{Chance top-1} & 5.6 & 5.8 & 5.8 & 5.9 & 6.3 & 6.5 & 6.7 & 7.2 & 7.3 & 7.8 & 7.8 & 8.0 & 8.0 & 8.0 \\
 
\bottomrule
\end{tabular}
\tablenotes{2013--2024: 500 random US calls a year with at least three exchanges. 2025 and 2026: every call of the main test in that year. TF-IDF here uses word weights fitted on calls up to 2012. The stage-1 checkpoint was chosen on calls from 2013--2015, and the 2023 models on calls from 2024, so these years are not untouched tests for those models. \emph{Chance top-1} is the mean of one over the number of answers in the call.}
\end{table}

\begin{table}[p]
\centering
\scriptsize
\caption{Within-call answer retrieval on the 2025--2026 test: every reader.}
\label{tab:retrieval-full}
\setlength{\tabcolsep}{4pt}\renewcommand{\arraystretch}{0.94}
\begin{tabular}{@{}L{6.2cm}rcrrrl@{}}
\toprule
Reader & Top-1 & 95\% interval & \makecell[c]{Top-1,\\call-weighted} & Top-3 & MRR & $\Delta$ to released [95\% interval] \\
\midrule
\multicolumn{7}{@{}l}{\textit{Adapted on calls up to 2023}} \\
\quad \textbf{NoLBERT-Calls-2023 (released)} & 78.0 & [77.8, 78.2] & 80.2 & 94.0 & 86.4 & \textemdash \\
\qquad + word overlap & 79.0 & [78.8, 79.2] & 81.0 & 94.4 & 87.1 & $-$1.0 [$-$1.0, $-$0.9] \\
\quad ChronoBERT-Calls-2023 at 256 tokens & 77.0 & [76.8, 77.2] & 79.2 & 93.5 & 85.7 & 1.0 [0.9, 1.2] \\
\qquad + word overlap & 77.9 & [77.7, 78.1] & 80.0 & 93.9 & 86.3 & 0.1 [$-$0.0, 0.3] \\
\quad FinBERT-Calls-2023 & 74.7 & [74.5, 74.9] & 77.0 & 92.8 & 84.2 & 3.4 [3.2, 3.5] \\
\qquad + word overlap & 77.1 & [76.9, 77.3] & 79.2 & 94.0 & 85.9 & 0.9 [0.7, 1.1] \\
\quad ChronoBERT-Calls-2023 & 72.3 & [72.1, 72.6] & 74.7 & 91.2 & 82.4 & 5.7 [5.5, 5.9] \\
\qquad + word overlap & 74.6 & [74.4, 74.9] & 76.9 & 92.5 & 84.1 & 3.4 [3.2, 3.6] \\
\quad DeBERTa-Calls-2023 & 68.2 & [68.0, 68.4] & 71.1 & 89.6 & 79.7 & 9.8 [9.6, 10.0] \\
\qquad + word overlap & 74.4 & [74.2, 74.6] & 76.7 & 92.5 & 84.0 & 3.6 [3.5, 3.8] \\
\quad NoLBERT-Calls-2023, same recipe & 67.0 & [66.7, 67.2] & 69.6 & 88.1 & 78.6 & 11.1 [10.9, 11.2] \\
\qquad + word overlap & 70.6 & [70.3, 70.8] & 73.1 & 90.5 & 81.3 & 7.5 [7.3, 7.6] \\
\addlinespace[2pt]
\multicolumn{7}{@{}l}{\textit{Adapted on calls up to 2012}} \\
\quad NoLBERT-Calls-2012, stage 1 & 71.4 & [71.2, 71.6] & 73.8 & 90.6 & 81.7 & 6.7 [6.5, 6.8] \\
\qquad + word overlap & 73.9 & [73.7, 74.1] & 76.3 & 92.4 & 83.7 & 4.1 [4.0, 4.3] \\
\quad ChronoBERT-Calls-2012 & 69.3 & [69.1, 69.5] & 71.8 & 89.5 & 80.3 & 8.8 [8.6, 8.9] \\
\qquad + word overlap & 72.5 & [72.3, 72.7] & 74.9 & 91.7 & 82.7 & 5.5 [5.4, 5.7] \\
\quad NoLBERT-Calls-2012, same recipe & 65.1 & [64.9, 65.3] & 67.9 & 86.8 & 77.1 & 12.9 [12.8, 13.1] \\
\qquad + word overlap & 69.5 & [69.3, 69.7] & 72.1 & 89.8 & 80.5 & 8.6 [8.4, 8.7] \\
\addlinespace[2pt]
\multicolumn{7}{@{}l}{\textit{Adapted on calls up to 2012, other}} \\
\quad NoLBERT-Calls-2012, 150-minute stage 1 & 60.6 & [60.3, 60.8] & 63.5 & 83.5 & 73.6 & 17.5 [17.3, 17.7] \\
\qquad + word overlap & 66.4 & [66.2, 66.6] & 69.0 & 87.8 & 78.1 & 11.6 [11.5, 11.8] \\
\addlinespace[2pt]
\multicolumn{7}{@{}l}{\textit{Stage-2 development runs (not released)}} \\
\quad v1: warm start & 73.8 & [73.5, 74.0] & 76.1 & 92.0 & 83.5 & 4.3 [4.1, 4.4] \\
\qquad + word overlap & 75.8 & [75.6, 76.0] & 78.0 & 93.3 & 85.0 & 2.2 [2.0, 2.3] \\
\quad v2: + same-call packing & 74.4 & [74.2, 74.6] & 76.7 & 92.3 & 83.9 & 3.6 [3.5, 3.7] \\
\qquad + word overlap & 76.6 & [76.4, 76.8] & 78.7 & 93.5 & 85.5 & 1.5 [1.3, 1.6] \\
\quad v3: v2 + QA-only batches, lr $3\times10^{-5}$ & 75.8 & [75.6, 76.0] & 78.0 & 93.1 & 84.9 & 2.2 [2.1, 2.3] \\
\qquad + word overlap & 77.6 & [77.4, 77.8] & 79.7 & 94.2 & 86.3 & 0.4 [0.2, 0.5] \\
\quad v4: v2 + 256-token answers & 77.7 & [77.5, 77.9] & 79.9 & 93.9 & 86.2 & 0.3 [0.2, 0.4] \\
\qquad + word overlap & 78.6 & [78.4, 78.8] & 80.7 & 94.5 & 86.9 & $-$0.5 [$-$0.7, $-$0.4] \\
\quad Ensemble of v4 and v3 & 78.4 & [78.2, 78.6] & 80.5 & 94.3 & 86.7 & $-$0.4 [$-$0.5, $-$0.3] \\
\qquad + word overlap & 79.3 & [79.1, 79.5] & 81.3 & 94.9 & 87.4 & $-$1.2 [$-$1.4, $-$1.1] \\
\addlinespace[2pt]
\multicolumn{7}{@{}l}{\textit{Read at 256 tokens without retraining}} \\
\quad FinBERT-Calls-2023 & 74.1 & [73.9, 74.4] & 76.5 & 92.9 & 84.0 & 3.9 [3.7, 4.1] \\
\qquad + word overlap & 76.6 & [76.4, 76.8] & 78.7 & 94.0 & 85.7 & 1.4 [1.3, 1.6] \\
\addlinespace[2pt]
\multicolumn{7}{@{}l}{\textit{Off-the-shelf retrievers}} \\
\quad E5-base-v2 & 59.3 & [59.1, 59.5] & 62.2 & 82.4 & 72.5 & 18.7 [18.5, 18.9] \\
\qquad + word overlap & 65.3 & [65.1, 65.6] & 68.0 & 87.0 & 77.3 & 12.7 [12.5, 12.9] \\
\quad BGE-base-en-v1.5 & 58.1 & [57.9, 58.3] & 61.0 & 80.8 & 71.3 & 19.9 [19.7, 20.2] \\
\qquad + word overlap & 64.0 & [63.8, 64.2] & 66.8 & 85.9 & 76.2 & 14.0 [13.8, 14.3] \\
\quad all-mpnet-base-v2 & 57.5 & [57.2, 57.7] & 60.6 & 81.4 & 71.2 & 20.6 [20.4, 20.8] \\
\qquad + word overlap & 64.8 & [64.6, 65.0] & 67.5 & 86.8 & 77.0 & 13.2 [13.0, 13.4] \\
\addlinespace[2pt]
\multicolumn{7}{@{}l}{\textit{Word overlap}} \\
\quad BM25 & 58.2 & [58.0, 58.5] & 61.1 & 80.7 & 71.2 & 19.8 [19.6, 20.0] \\
\quad TF-IDF & 57.3 & [57.0, 57.5] & 60.2 & 80.4 & 70.6 & 20.8 [20.6, 21.0] \\
\addlinespace[2pt]
\multicolumn{7}{@{}l}{\textit{Pretrained encoders, zero-shot}} \\
\quad FinBERT & 31.9 & [31.7, 32.1] & 35.5 & 59.1 & 49.9 & 46.2 [45.9, 46.4] \\
\qquad + word overlap & 59.4 & [59.2, 59.7] & 62.3 & 82.5 & 72.6 & 18.6 [18.4, 18.8] \\
\quad NoLBERT & 26.3 & [26.1, 26.5] & 29.8 & 51.7 & 44.4 & 51.7 [51.5, 51.9] \\
\qquad + word overlap & 58.6 & [58.4, 58.9] & 61.6 & 81.7 & 71.9 & 19.4 [19.2, 19.6] \\
\quad ChronoBERT-2023 & 24.1 & [23.9, 24.2] & 27.3 & 48.1 & 41.8 & 54.0 [53.7, 54.2] \\
\qquad + word overlap & 58.2 & [58.0, 58.5] & 61.1 & 81.3 & 71.5 & 19.8 [19.6, 20.0] \\
\quad ChronoBERT-2012 & 23.7 & [23.6, 23.9] & 26.9 & 47.3 & 41.3 & 54.3 [54.1, 54.5] \\
\qquad + word overlap & 58.2 & [57.9, 58.4] & 61.2 & 81.4 & 71.5 & 19.9 [19.7, 20.1] \\
\quad DeBERTa-v3-base & 10.0 & [9.9, 10.1] & 12.5 & 27.9 & 26.9 & 68.0 [67.8, 68.2] \\
\qquad + word overlap & 57.3 & [57.1, 57.6] & 60.3 & 80.7 & 70.8 & 20.7 [20.5, 20.9] \\
 
\bottomrule
\end{tabular}
\tablenotes{Percent; intervals from bootstrap resamples of calls. \emph{Call-weighted} averages top-1 within each call first. $\Delta$ is the released model's top-1 without word overlap minus the row's top-1, in points, with a paired interval; a negative value favors the row. \emph{+ word overlap} rows blend the reader with TF-IDF using the weight chosen on the validation calls.}
\end{table}

\begin{table}[!htbp]
\centering
\scriptsize
\caption{Stage-2 runs. Every run lasts 150 minutes on one T4 and starts from stage~1, except v4c, which continues v4.}
\label{tab:development}
\setlength{\tabcolsep}{3.4pt}
\begin{tabular}{@{}lL{4.1cm}ccccrrrr@{}}
\toprule
Run & Change & \makecell[c]{Exchanges\\per call} & \makecell[c]{QA\\batches} & \makecell[c]{Learning\\rate} & \makecell[c]{Answer\\tokens} & \makecell[c]{Steps\\(best)} & \makecell[c]{Validation\\top-1} & \makecell[c]{Test\\top-1} & \makecell[c]{Test,\\+ word\\overlap} \\
\midrule
v1 & Warm start from stage 1 & 8 & 75\% & $2\times10^{-5}$ & 128 & 6,963 (6,270) & 75.8 & 73.8 & 75.8 \\
v2 & + same-call packing & 32 & 75\% & $2\times10^{-5}$ & 128 & 7,239 (7,239) & 77.1 & 74.4 & 76.6 \\
v3 & v2 + QA-only batches, higher LR & 32 & 100\% & $3\times10^{-5}$ & 128 & 6,882 (4,821) & 78.5 & 75.8 & 77.6 \\
v4 & v2 + 256-token answers & 32 & 75\% & $2\times10^{-5}$ & 256 & 3,397 (3,397) & 80.3 & 77.7 & 78.6 \\
\rowcolor{softteal}
\textbf{v4c (released)} & v4 + a second 150-minute run & 32 & 75\% & $2\times10^{-5}$ & 256 & 3,722 (3,722) & 80.7 & 78.0 & 79.0 \\
Ensemble & Average of v4 and v3 (not released) & & & & & & 80.3 & 78.4 & 79.3 \\
 
\bottomrule
\end{tabular}
\tablenotes{Validation: call-weighted top-1 on the 300 calls of 2024, the basis of every choice. Test: question-weighted top-1 on 2025--2026. \emph{QA batches}: share of batches made of question--answer pairs; the rest pair passages of the same call. The ensemble averages the standardized similarities of v4 and v3.}
\end{table}

\begin{table}[!htbp]
\centering
\scriptsize
\caption{Tasks on frozen representations: every reader.}
\label{tab:tasks-full}
\setlength{\tabcolsep}{4pt}
\begin{tabular}{@{}L{5.4cm}rrrrrr@{}}
\toprule
 & \multicolumn{2}{c}{Passage matching} & & \multicolumn{2}{c}{Masked terms} & \\
\cmidrule(lr){2-3}\cmidrule(lr){5-6}
Reader & Top-1 & Top-10 & Role & Top-1 & Top-5 & \makecell[c]{SubjECTive-QA\\probe} \\
\midrule
\rowcolor{softteal}
\textbf{NoLBERT-Calls-2023 (released)} & 7.6 & 21.9 & 96.2 & 25.6 & 47.1 & 64.8 \\
NoLBERT-Calls-2012, stage 1 & 5.8 & 17.9 & 95.9 & 23.9 & 45.1 & 65.4 \\
NoLBERT-Calls-2023, same recipe & 6.3 & 15.8 & 95.6 & 21.3 & 40.2 & 65.1 \\
v4 (stage-2 run 1) & \topB{8.3} & \topC{22.1} & 96.6 & 24.9 & 46.4 & 65.2 \\
v3 & 3.9 & 11.5 & 95.6 & \topC{25.9} & \topC{48.1} & 65.0 \\
NoLBERT-Calls-2012, 150-minute stage 1 & 4.0 & 12.8 & 96.0 & 19.4 & 36.9 & 65.2 \\
NoLBERT-Calls-2012, same recipe & 6.8 & 15.5 & 95.7 & 21.9 & 40.1 & 65.2 \\
FinBERT-Calls-2023 & \topC{8.2} & \topA{27.9} & \topB{97.8} & \topA{39.6} & \topA{65.0} & \topA{66.5} \\
ChronoBERT-Calls-2023 & 6.6 & 19.2 & 95.4 & 25.2 & 47.1 & 65.4 \\
DeBERTa-Calls-2023 & 2.7 & 15.2 & \topC{97.2} & 20.4 & 41.2 & 65.0 \\
ChronoBERT-Calls-2012 & 5.6 & 17.9 & 95.0 & 25.1 & 46.6 & 65.0 \\
E5-base-v2 & 5.2 & 14.0 & 96.5 & \textemdash & \textemdash & \topC{66.2} \\
TF-IDF & \topA{12.6} & \topB{26.0} & 88.1 & \textemdash & \textemdash & \topB{66.3} \\
BGE-base-en-v1.5 & 5.4 & 15.1 & 93.7 & \textemdash & \textemdash & 65.9 \\
all-mpnet-base-v2 & 6.5 & 14.4 & 94.1 & \textemdash & \textemdash & 65.4 \\
NoLBERT, zero-shot & 0.8 & 4.1 & 96.6 & 18.7 & 34.6 & 65.3 \\
FinBERT, zero-shot & 0.9 & 6.9 & \topA{98.2} & \topB{38.0} & \topB{62.6} & 66.0 \\
ChronoBERT-2023, zero-shot & 0.5 & 3.1 & 96.1 & 23.6 & 43.5 & 64.8 \\
ChronoBERT-2012, zero-shot & 0.7 & 3.8 & 96.0 & 21.8 & 40.4 & 64.4 \\
DeBERTa-v3-base, zero-shot & 0.1 & 1.2 & 90.2 & 0.0 & 0.0 & 63.2 \\
\addlinespace[2pt]
\textit{Chance or majority class} & 0.1 & 1.0 & 50.0 & & & 55.6 \\
 
\bottomrule
\end{tabular}
\tablenotes{As in Table~\ref{tab:tasks}. A dash means the task is not defined for the reader. \shadekey.}
\end{table}

\begin{table}[!htbp]
\centering
\scriptsize
\caption{SubjECTive-QA after fine-tuning, property by property: weighted F1, mean (standard deviation) over the authors' three splits.}
\label{tab:sqa-props}
\setlength{\tabcolsep}{3.4pt}
\begin{tabular}{@{}L{4.4cm}ccccccc@{}}
\toprule
Encoder & Assertive & Cautious & Clear & Optimistic & Relevant & Specific & Mean \\
\midrule
NoLBERT, zero-shot base & 49.0 (1.4) & 59.9 (2.2) & 82.5 (0.8) & 55.4 (0.9) & 84.9 (1.1) & 54.8 (1.6) & 64.4 (0.9) \\
NoLBERT-Calls-2023, same recipe & 49.9 (3.6) & 61.0 (2.3) & 82.5 (0.8) & 59.5 (1.7) & 84.8 (1.1) & 56.6 (1.3) & 65.7 (1.0) \\
v3 & 49.9 (4.0) & 61.2 (2.7) & 82.5 (0.8) & 60.0 (1.2) & 84.9 (1.1) & 56.4 (2.5) & 65.8 (1.4) \\
\rowcolor{softteal}
\textbf{NoLBERT-Calls-2023 (released)} & 50.3 (4.3) & 61.3 (3.0) & 82.5 (0.8) & 59.4 (2.8) & 84.9 (1.1) & 56.3 (2.8) & 65.8 (1.5) \\
FinBERT-Calls-2023 & 50.0 (4.3) & 62.0 (1.9) & 82.4 (0.7) & 58.2 (3.6) & 85.2 (1.1) & 59.4 (1.9) & 66.2 (0.9) \\
ChronoBERT-Calls-2023 & 48.8 (4.1) & 60.7 (1.1) & 82.8 (1.2) & 60.0 (1.4) & 84.8 (1.0) & 56.7 (0.3) & 65.6 (1.0) \\
DeBERTa-Calls-2023 & 50.9 (3.2) & 59.8 (2.1) & 82.5 (0.8) & 59.7 (4.3) & 84.9 (1.1) & 55.2 (2.3) & 65.5 (1.5) \\
\addlinespace[2pt]
\textit{Released model, macro F1} & 37.5 (2.5) & 31.7 (3.7) & 31.2 (0.1) & 41.7 (3.0) & 31.5 (0.1) & 47.5 (3.7) & 36.8 (1.2) \\
 
\bottomrule
\end{tabular}
\tablenotes{Clear and Relevant give the same weighted F1 for almost every encoder because almost every head predicts the majority class; the macro F1 of the released model on these two properties, about 31, confirms it. The last column averages the six properties within each split and reports the mean and standard deviation of the three split averages.}
\end{table}

\clearpage
\section{Training Details}
\label{app:training}

Table~\ref{tab:hyper} lists the settings of the two stages and Table~\ref{tab:compute} the compute behind every model in the paper.

\begin{table}[!htbp]
\centering
\footnotesize
\caption{Settings of the two adaptation stages.}
\label{tab:hyper}
\setlength{\tabcolsep}{6pt}
\begin{tabular}{@{}L{4.4cm}L{5.1cm}L{5.1cm}@{}}
\toprule
Setting & Stage 1 & Stage 2 (released) \\
\midrule
Starting weights & NoLBERT & Selected stage 1 checkpoint \\
Training calls & All countries, up to 31 Dec 2012 & US, up to 31 Dec 2023 \\
Batches & 75\% question--answer, 25\% passage pairs & 75\% question--answer, 25\% passage pairs \\
Exchanges of one call per batch & Up to 8 & Up to 32 \\
Tokens, question and answer & 128 and 128 & 128 and 256 \\
Batch size & 32 & 32 \\
Temperature $\tau$ & 0.05 & 0.05 \\
Masked-word loss & 15\% masking, weight $\lambda = 1$ & 15\% masking, weight $\lambda = 1$ \\
Optimizer & AdamW, learning rate $2\times10^{-5}$, weight decay 0.01, betas (0.9, 0.98), clipping at 1 & Same \\
Schedule & 5\% warm-up, then cosine decay & Same, restarted for the second run \\
Precision & bfloat16 & float16 with gradient scaling and checkpointing \\
Budget & 480 minutes on one RTX 3060 & Two runs of 150 minutes on one T4 \\
Steps & About 19,000 & 3,397 and 3,722 \\
Validation calls & 300 calls from 2013--2015 & 300 US calls from 2024 \\
Validation metric & Top-1 pooled over questions, every 15 minutes & Call-weighted top-1, every 15 minutes \\
Selected checkpoint & Step 17,451, validation top-1 70.3\% & Last step of run 2, validation top-1 80.7\% \\
 
\bottomrule
\end{tabular}
\end{table}

\begin{table}[!htbp]
\centering
\footnotesize
\caption{Compute. Evaluation runs are not included.}
\label{tab:compute}
\setlength{\tabcolsep}{6pt}
\begin{tabular}{@{}L{6.4cm}L{3.6cm}L{4.4cm}@{}}
\toprule
Item & Hardware & Budget \\
\midrule
Stage 1 & One RTX 3060 (12 GB) & 480 minutes \\
Stage 2, released model & One T4 per run & 2 $\times$ 150 minutes \\
\rowcolor{softteal}
\textbf{Released model, total} & & \textbf{About 13 GPU-hours} \\
\addlinespace
Stage 2 development runs (v1--v3) & One T4 each & 3 $\times$ 150 minutes \\
Same-recipe adaptations, six models & One T4 each & 6 $\times$ 150 minutes \\
Stage 1, 150-minute run (reference) & One RTX 3060 & 150 minutes \\
 
\bottomrule
\end{tabular}
\end{table}

\end{document}